\documentclass[10pt,journal,compsoc]{IEEEtran}

\usepackage{graphicx}
\usepackage{comment}
\usepackage{amsmath,amssymb}
\usepackage{color}
\usepackage{url}
\usepackage{booktabs, multirow, array, makecell, caption}
\usepackage{verbatimbox}
\usepackage{lipsum}
\usepackage{cuted}
\usepackage{stfloats}
\usepackage{bigstrut}
\usepackage{hyperref}
\usepackage[table]{xcolor}
\usepackage{rotating}
\usepackage{bm}

\ifCLASSOPTIONcompsoc
  \usepackage[nocompress]{cite}
\else
  \usepackage{cite}
\fi

\begin{document}

\title{Polarimetric Multi-View Inverse Rendering}

\author{Jinyu~Zhao,
   Yusuke~Monno,~\IEEEmembership{Member,~IEEE,}
   and~Masatoshi~Okutomi,~\IEEEmembership{Member,~IEEE}%
   \IEEEcompsocitemizethanks{\IEEEcompsocthanksitem The authors are with the Department of System and Control Engineering, Tokyo Institute of Technology, 2-12-1 Ookayama, Meguro-ku, Tokyo, Japan 152-8550.\protect\\
   E-mail: jzhao@ok.sc.e.titech.ac.jp}
   }

\markboth{IEEE TRANSACTIONS ON PATTERN ANALYSIS AND MACHINE INTELLIGENCE}
{Shell \MakeLowercase{\textit{et al.}}: Bare Demo of IEEEtran.cls for Computer Society Journals}

\IEEEtitleabstractindextext{%
   \begin{abstract}
      A polarization camera has great potential for 3D reconstruction since the angle of polarization~(AoP) and the degree of polarization~(DoP) of reflected light are related to an object's surface normal. In this paper, we propose a novel 3D reconstruction method called Polarimetric Multi-View Inverse Rendering~(Polarimetric MVIR) that effectively exploits geometric, photometric, and polarimetric cues extracted from input multi-view color-polarization images. We first estimate camera poses and an initial 3D model by geometric reconstruction with a standard structure-from-motion and multi-view stereo pipeline. We then refine the initial model by optimizing photometric rendering errors and polarimetric errors using multi-view RGB, AoP, and DoP images, where we propose a novel polarimetric cost function that enables an effective constraint on the estimated surface normal of each vertex, while considering four possible ambiguous azimuth angles revealed from the AoP measurement. The weight for the polarimetric cost is effectively determined based on the DoP measurement, which is regarded as the reliability of polarimetric information. Experimental results using both synthetic and real data demonstrate that our Polarimetric MVIR can reconstruct a detailed 3D shape without assuming a specific surface material and lighting condition.
   \end{abstract}

   }

\maketitle
\IEEEdisplaynontitleabstractindextext
\IEEEpeerreviewmaketitle

\IEEEraisesectionheading{\section{Introduction}\label{sec:introduction}}

\IEEEPARstart{I}{mage-based} 3D reconstruction has been studied for years and can be applied to various applications, e.g. model creation~\cite{biehler20143d}, localization~\cite{cao2013graph}, segmentation~\cite{dai20183dmv},
and shape recognition~\cite{su2015multi}. There are two common approaches for 3D reconstruction: geometric reconstruction and photometric reconstruction. The geometric reconstruction is based on feature extraction, matching, and triangulation using multi-view images. It has been well established as structure from motion~(SfM)~\cite{agarwal2009building,schonberger2016structure,wu2011high} for sparse point cloud reconstruction, which is often followed by dense reconstruction with multi-view stereo~(MVS)~\cite{furukawa2010towards,furukawa2009accurate,galliani2015massively}. On the other hand, the photometric reconstruction exploits shading information for each image pixel to derive dense surface normals. It has been well studied as shape from shading~\cite{barron2014shape,xiong2014shading,zhang1999shape} and photometric stereo~\cite{haefner2019variational,ikehata2014photometric,wu2010robust}.

There also exist some advanced methods combining the advantages of both approaches, e.g. multi-view photometric stereo~\cite{li2020multi,park2016robust} and multi-view inverse rendering~(MVIR)~\cite{kim2016multi,li2021spectral}. These methods typically start from the geometric reconstruction with SfM and MVS for camera pose estimation and initial model reconstruction, and then refine the initial model, especially for texture-less surfaces, by utilizing photometric shading cues.

\begin{figure}[t!]
   \centering
   \includegraphics[trim={0cm 0cm 0cm 0cm}, width=1.0\linewidth]{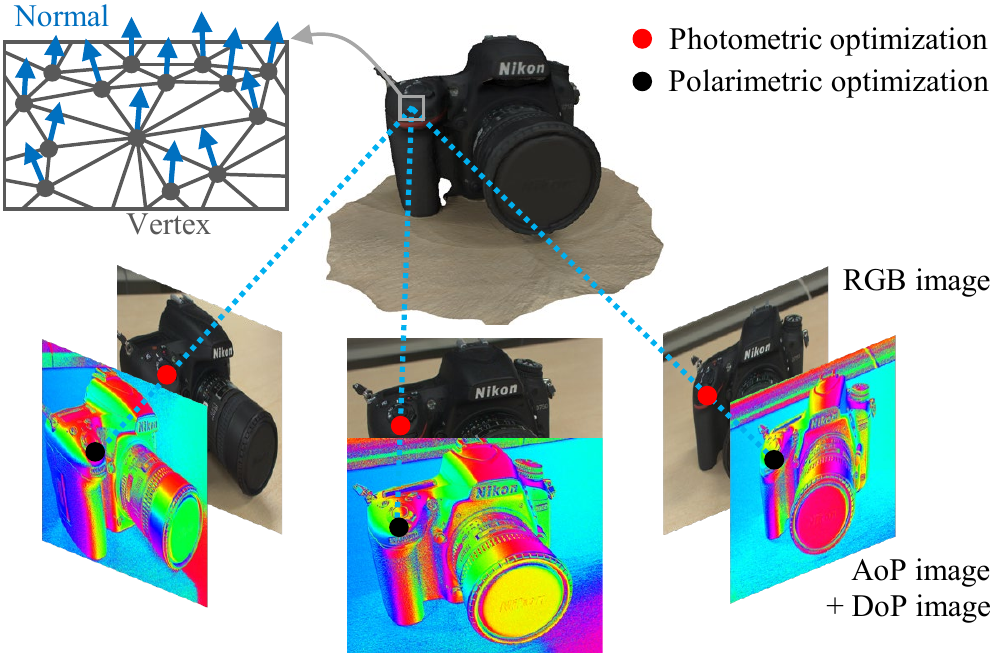}
   \caption{Overview of our Polarimetric MVIR framework: Using estimated camera poses and an initial model from SfM and MVS, Polarimetric MVIR refines the initial model by optimizing photometric rendering errors and polarimetric errors by using multi-view RGB, AoP, and DoP images, where each vertex normal is constrained based on polarimetric information.}
   \label{fig:overall2}
\end{figure}

Multi-view reconstruction using polarization images \cite{cui2017polarimetric,yang2018polarimetric} has also received increasing attention with the development of one-shot polarization cameras using Sony IMX250 monochrome- or color-polarization sensor~\cite{maruyama20183}, e.g. JAI GO-5100MP-PGE~\cite{JAI} and Lucid PHX050S-Q~\cite{Lucid} cameras. The use of polarimetric information has great potential for 3D reconstruction since the angle of polarization~(AoP) and the degree of polarization~(DoP) of reflected light are related to the azimuth and the zenith angles of the object's surface normal, respectively. One state-of-the-art method is Polarimetric MVS~\cite{cui2017polarimetric}, which generates dense depth maps for each view by propagating initial sparse depths from SfM by using the AoP image of the corresponding view obtained using a polarization camera.
Since there are four possible azimuth angles corresponding to one AoP measurement as will be detailed in Section~\ref{sec:p-ambiguities}, the depth propagation relies on the disambiguation of these polarimetric ambiguities using the initial sparse depth cues from SfM.

In this paper, inspired by the successes of MVIR~\cite{kim2016multi} and Polarimetric MVS~\cite{cui2017polarimetric}, we propose Polarimetric Multi-View Inverse Rendering (Polarimetric MVIR), which is a fully passive 3D reconstruction method exploiting all geometric, photometric, and polarimetric cues. 
Figure~\ref{fig:overall2} illustrates our Polarimetric MVIR framework.
We first estimate camera poses and an initial surface model based on SfM and MVS. We then refine the initial model by simultaneously using multi-view RGB, AoP, and DoP images while estimating the positions of surface vertices, vertex albedos, and illuminations for each image. The key of our method is a novel global cost optimization framework for shape refinement. 
In addition to a standard photometric rendering term that evaluates RGB rendering errors, we introduce a novel polarimetric term that evaluates the difference between the azimuth angle of each estimated surface vertex's normal and four possible azimuth angles revealed from the corresponding AoP measurement.
Our polarimetric term enables us to take all four possible ambiguous azimuth angles into account in the global optimization, 
without the necessity of explicitly solving the ambiguity, which makes our method robust to mis-disambiguation. 
The weight for the polarimetric term is effectively determined by the DoP measurement, which indicates the strength of polarization and thus can be used as a reliability metric for our polarimetric term.
Experimental results using synthetic and real data demonstrate that, compared with existing MVS, MVIR, and Polarimetric MVS, our Polarimetric MVIR can reconstruct a more detailed 3D model from unconstrained input images without any prerequisites for the surface material and the lighting condition.

\begin{table}[t!]
   \centering
   \renewcommand\arraystretch{1.2}
   \setlength\aboverulesep{0pt}\setlength\belowrulesep{0pt}
   \setcellgapes{1.5pt}\makegapedcells
   \caption{Types of information used by related methods}
   \label{table:position}
   \scalebox{0.95}{
\begin{tabular}{l|l|p{4.8em}|p{4.8em}}
   \toprule
         & \multicolumn{1}{p{4.8em}|}{Geometry} & Photometry & Polarimetry \\
   \midrule
   \midrule
   MVIR~\cite{kim2016multi} & \multicolumn{1}{p{4.8em}|}{\checkmark} & \checkmark &  \\
   \midrule
   Polarimetric MVS~\cite{cui2017polarimetric} & \multicolumn{1}{p{4.8em}|}{\checkmark} & \multicolumn{1}{l|}{} & \checkmark \\
   \midrule
   Polarimetric MVIR (Ours) &   \checkmark    & \checkmark & \checkmark \\
   \bottomrule
   \end{tabular}%  
   }
 \end{table}%
 
Table~\ref{table:position} summarizes the properties of used information in closely related works: MVIR, Polarimetric MVS, and our Polarimetric MVIR. All the methods apply SfM and MVS as initial geometric reconstruction. MVIR then optimizes geometry and photometry information for estimating a fine shape as well as albedo and illumination parameters, while Polarimetric MVS applies geometry and polarimetry information to reconstruct dense depth maps for every view.
Compared to previous works, main contributions of our work are summarized as below.
\begin{itemize}
   \item We propose Polarimetric MVIR, which is the first 3D reconstruction method based on multi-view geometric, photometric, and polarimetric information with an inverse rendering framework. 
   \item We propose a novel polarimetric cost function that enables us to effectively constrain the normal of each estimated vertex while solving the azimuth angle ambiguities as a global optimization problem.
   \item We experimentally validate the robustness of our method to polarimetric ambiguities, which makes our method unconstrained to a specific material and lighting condition.
\end{itemize}

A preliminary version of this work was presented earlier in~\cite{zhao2020polarimetric}. In this extended work, we have newly incorporated the DoP measurement to determine the balancing weights between the polarimetric cost term and the other cost terms, which contributes to performance improvements by adopting only reliable polarimetric information to our framework.
We have also newly conducted experiments using Mitsuba~2 renderer~\cite{nimier2019mitsuba}, which can simulate realistic polarization images based on a polarimetric bidirectional reflectance distribution function (pBRDF)~\cite{baek2018simultaneous}. By using Mitsuba~2 renderer, we have added various simulation results to quantitatively validate our method in terms of the effectiveness of each proposed component, the accuracy of shape, albedo, and illumination estimation, and the robustness to polarimetric ambiguities caused by different materials.

\section{Related Works}
In the past literature, a number of methods have been proposed for the geometric 3D reconstruction~(e.g. SfM~\cite{agarwal2009building,schonberger2016structure,wu2011high} and MVS~\cite{furukawa2010towards,furukawa2009accurate,galliani2015massively}) and the photometric 3D reconstruction~(e.g. shape from shading~\cite{barron2014shape,xiong2014shading,zhang1999shape} and photometric stereo~\cite{haefner2019variational,ikehata2014photometric,wu2010robust}).
In this section, we briefly introduce the combined methods of geometric and photometric 3D reconstruction, and also polarimetric 3D reconstruction methods, which are closely related to our work.

\subsection{Multi-view geometric-photometric reconstruction}
The geometric 3D reconstruction approach generally begins with SfM, from which camera poses and a sparse point cloud are derived by matching feature points among multi-view images. Then, MVS is typically applied to obtain a denser point cloud by exploiting epipolar constraints from estimated camera poses. 
The geometric approach is relatively robust to estimate camera poses and a sparse or dense point cloud for well-textured surfaces, owing to robust feature detection and matching algorithms~\cite{bay2008speeded,lowe2004distinctive}.
However, it is weak in texture-less surfaces because sufficient feature correspondences cannot be obtained for those surfaces. In contrast, the photometric 3D reconstruction approach recovers fine details for texture-less surfaces by exploiting pixel-by-pixel shading information. However, it generally assumes a known or calibrated camera and lighting setup.

To take the advantages of both approaches, other advanced methods~\cite{maurer2016combining,wu2010fusing,wu2011high,li2020multi,park2016robust,kim2016multi}, such as multi-view photometric stereo~\cite{li2020multi,park2016robust} and MVIR~\cite{kim2016multi}, combine the two approaches.
These methods typically estimate camera poses and an initial coarse model based on SfM and MVS
and then refine the initial model, especially in texture-less regions, by using shading cues from multiple viewpoints. To remove the necessity of a known or calibrated lighting setup, MVIR~\cite{kim2016multi} applies general illumination models and jointly estimates a refined shape, spatially-varying surface albedos, and each image's illumination. Our Polarimetric MVIR is built on this MVIR approach to realize a fully passive and uncalibrated method.

\subsection{Single-view shape from polarization~(SfP)}
There are many SfP methods which estimate object's surface normals~\cite{atkinson2006recovery,huynh2013shape,kadambi2015polarized,miyazaki2003polarization,morel2005polarization,smith2018height,tozza2017linear,ichikawa2021shape} based on the physical properties that AoP and DoP of reflected light are related to the azimuth and the zenith angles of the object's surface normal, respectively.
However, existing SfP methods usually assume a specific surface material because of the material-dependent ambiguous relationship between AoP and the azimuth angle, and also the ambiguous relationship between DoP and the zenith angle. 
For instance, a diffuse polarization model is adopted in~\cite{atkinson2006recovery,huynh2013shape,miyazaki2003polarization,tozza2017linear}, a specular polarization model is applied in~\cite{morel2005polarization}, and a dielectric material is considered in~\cite{kadambi2015polarized,smith2018height}.

Since a photometric 3D reconstruction method, such as a shape from shading and a photometric stereo method, derives estimated surface normals for each pixel, it can be used to provide cues to resolve the ambiguities between the polarimetric information and the surface normal. Based on this, some SfP methods exploit photometric information as an augment to resolve the polarimetric ambiguities.
However, these methods usually assume an uniform surface albedo and known light sources~\cite{atkinson2017polarisation,atkinson2007surface,ngo2015shape}. Although a recent method of~\cite{smith2018height} can handle spatially-varing albedos, it still assumes a distant point light source positioned in the same hemisphere as the viewer.
Therefore, existing single-view SfP methods are usually not passive and require the information about the lighting setup.

\subsection{Multi-view geometric-polarimetric reconstruction}
Multi-view geometry and polarization are also closely related and some studies have shown that multi-view polarimetric information is valuable for surface normal estimation~\cite{atkinson2007shape,ghosh2011multiview,miyazaki2020shape,miyazaki2016surface,rahmann2001reconstruction,fukao2021polarimetric}, camera pose estimation~\cite{chen2018polarimetric,cui2019polarimetric}, and light direction estimation~\cite{ngo2021surface}. 
However, to avoid the problem of polarimetric disambiguation, most of existing multi-view polarimetric 3D reconstruction methods make an assumption on the surface material, e.g. diffuse surfaces~\cite{atkinson2007shape,cui2019polarimetric,ngo2021surface}, specular surfaces~\cite{miyazaki2012polarization,miyazaki2016surface,rahmann2001reconstruction,miyazaki2020shape}, faces~\cite{ghosh2011multiview}, and transparent surfaces~\cite{miyazaki2004transparent}.

To relieve the constraint on the surface material, some methods utilize depth estimation as a reference to resolve the ambiguities, where an RGB-D sensor~\cite{kadambi2015polarized}, a stereo camera setup~\cite{zhu2019depth}, MVS~\cite{cui2017polarimetric} or SLAM~\cite{yang2018polarimetric}, is combined with the polarimetric reconstruction to provide depth cues for the polarimetric disambiguation.

Recent two state-of-the-art methods, Polarimetric MVS~\cite{cui2017polarimetric} and Polarimetric SLAM~\cite{yang2018polarimetric}, consider a mixed diffuse and specular reflection model to remove the necessity of known surface materials. These methods first disambiguate the ambiguity for AoP by using initial sparse depth cues from MVS or SLAM. Each viewpoint's depth map is then densified by propagating the sparse depth, where the disambiguated AoP values are used to find iso-depth contours along which the depth can be propagated. Although dense multi-view depth maps can be generated by the depth propagation, this approach relies on correct disambiguation which is not easy in general.

\subsection{Advantages of Polarimetric MVIR}
Compared to prior studies, our method has several advantages. First, it advances MVIR~\cite{kim2016multi} by using polarimetric information while inheriting the benefits of MVIR. Second, similar to~\cite{cui2017polarimetric,yang2018polarimetric}, our method is fully passive and does not require calibrated lighting and known surface materials. Third, polarimetric ambiguities are resolved as an optimization problem in shape refinement, instead of explicitly disambiguating them beforehand as in~\cite{cui2017polarimetric,yang2018polarimetric}, which can avoid relying on the assumption that the disambiguation is correct. Finally, a fine shape can be obtained by simultaneously exploiting photometric and polarimetric cues, where multi-view AoP measurements are used for constraining each estimated surface vertex's normal, which is a more direct and natural way to exploit azimuth-angle-related AoP measurements for shape estimation.

\section{Polarimetric Ambiguities in Surface Normal Prediction}
\label{sec:p-ambiguities}

\begin{figure}[t!]
   \centering
   \includegraphics[trim={0cm 0cm 0cm 0cm}, width=1.0\linewidth]{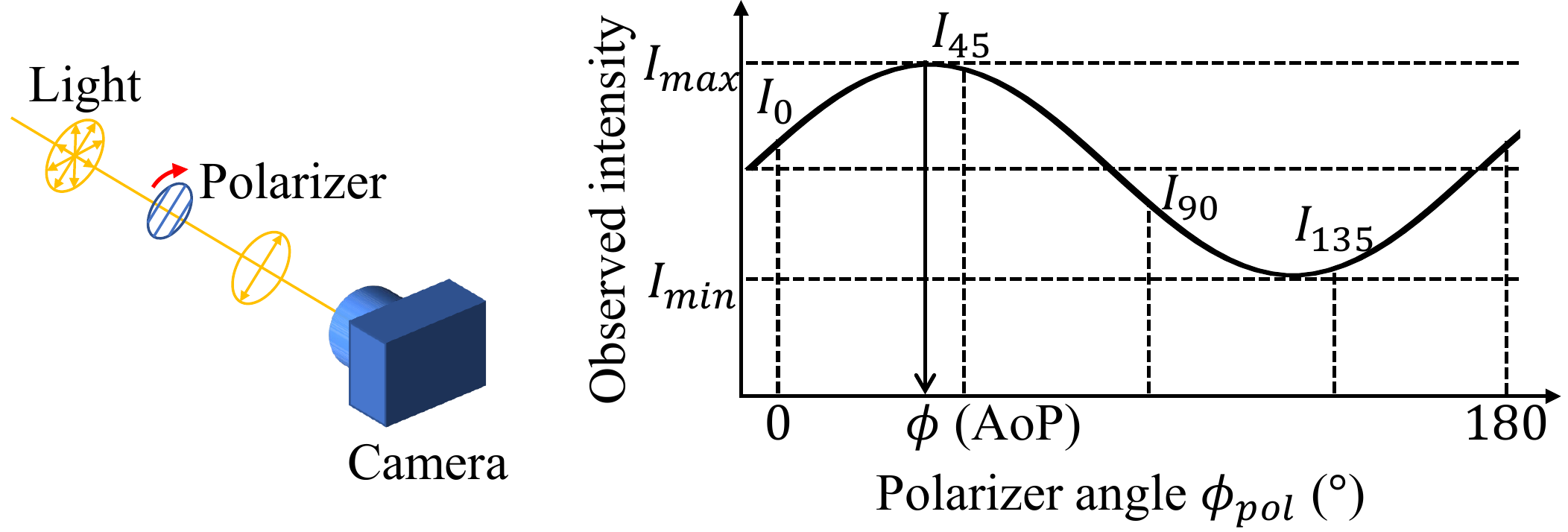}
   \caption{Observed intensity of light when rotating a polarizer in front of a camera (the orange arrows represent oscillations of the light's electric field): $I_{max}$ is observed when the polarizer angle equals AoP and $I_{min}$ is observed when the polarizer angle has $\pi/2$'s difference with AoP.}
   \label{fig:observation}
\end{figure}

\subsection{Polarimetric calculation}
\label{subsec:calculation}
Unpolarized light becomes partially polarized after reflection by a certain object's surface. Consequently, as shown in Fig.~\ref{fig:observation}, under common unpolarized illumination, the intensity of reflected light observed by a camera equipped with a polarizer satisfies the following equation:
\begin{equation}
   I_{\phi_{pol}}=\frac{I_{max}+I_{min}}{2}+\frac{I_{max}-I_{min}}{2}
   {\rm cos}2(\phi_{pol}-\phi),
\end{equation}
where $I_{max}$ and $I_{min}$
are the maximum and the minimum intensities observed when the polarizer is rotated around 180 degrees and $\phi$ is the reflected light's AoP, which indicates the strongest polarization direction. $I_{\phi_{pol}}$ is the observed intensity, where $\phi_{pol}$ is the polarizer angle at each capturing time. 
A polarization camera commonly observes the intensities of four polarization directions ($0^{\circ}$, $45^{\circ}$, $90^{\circ}$, and $135^{\circ}$), i.e. $I_0$, $I_{45}$, $I_{90}$, and $I_{135}$. 
From those measurements, Stokes vector~\cite{stokes1851composition} can be derived as
\begin{equation}
   \label{eq:Stokes}
   \textbf{s}=\left[\begin{matrix}
         s_0 \\s_1\\s_2\\s_3
      \end{matrix}
      \right]
   =\left[\begin{matrix}
         I_{max}+I_{min} \\(I_{max}-I_{min}){\rm cos}(2\phi)\\(I_{max}-I_{min}){\rm sin}(2\phi)\\0
      \end{matrix}
      \right]
   =\left[\begin{matrix}
         I_0+I_{90} \\I_0-I_{90}\\I_{45}-I_{135}\\0
      \end{matrix}
      \right],
\end{equation}
where $s_3=0$ because circularly polarized light is not considered in this work.
Then, AoP can be calculated using the Stokes vector as
\begin{equation}
   \label{eq:AoP}
   \phi=\frac{1}{2}{\rm tan}^{-1}\frac{s_2}{s_1}.
\end{equation}
In addition, DoP depicted as $\rho$, can also be computed using the Stokes vector as
   \begin{equation}
      \label{eq:DoP}
      \rho = \frac{I_{max}-I_{min}}{I_{max}+I_{min}} = \frac{\sqrt{s_1^2+s_2^2}}{s_0}.
   \end{equation}
According to Eq.~(\ref{eq:Stokes}) to (\ref{eq:DoP}), we can calculate the AoP and the DoP information from $I_0$, $I_{45}$, $I_{90}$, and $I_{135}$.

\subsection{Ambiguities}
\label{sec:ambiguities}
Under unpolarized incident light, AoP of reflected light reveals
information about the surface normal according to Fresnel equations,
as depicted by Atkinson and Hancock \cite{atkinson2006recovery}.
There are two linear polarization components of the incident wave:
$s$-polarized light and $p$-polarized light
whose directions of polarization are perpendicular and parallel to
the plane of incidence consisting of incident light and surface normal, respectively.

In this work, we consider a mixed polarization reflection model~\cite{baek2018simultaneous,cui2017polarimetric} under unpolarized lighting,
which includes unpolarized diffuse reflection, polarized specular reflection ($s$-polarized light is stronger),
and polarized diffuse reflection ($p$-polarized light is stronger).
In that case, the relationship between AoP and the azimuth angle, which is the angle between surface normal's projection to the image plane and $x$-axis in the image coordinates, depends on which polarized reflection's component is dominant.

\begin{figure}
   \centering
   \includegraphics[trim={0cm 0cm 0cm 0cm}, width=0.99\linewidth]{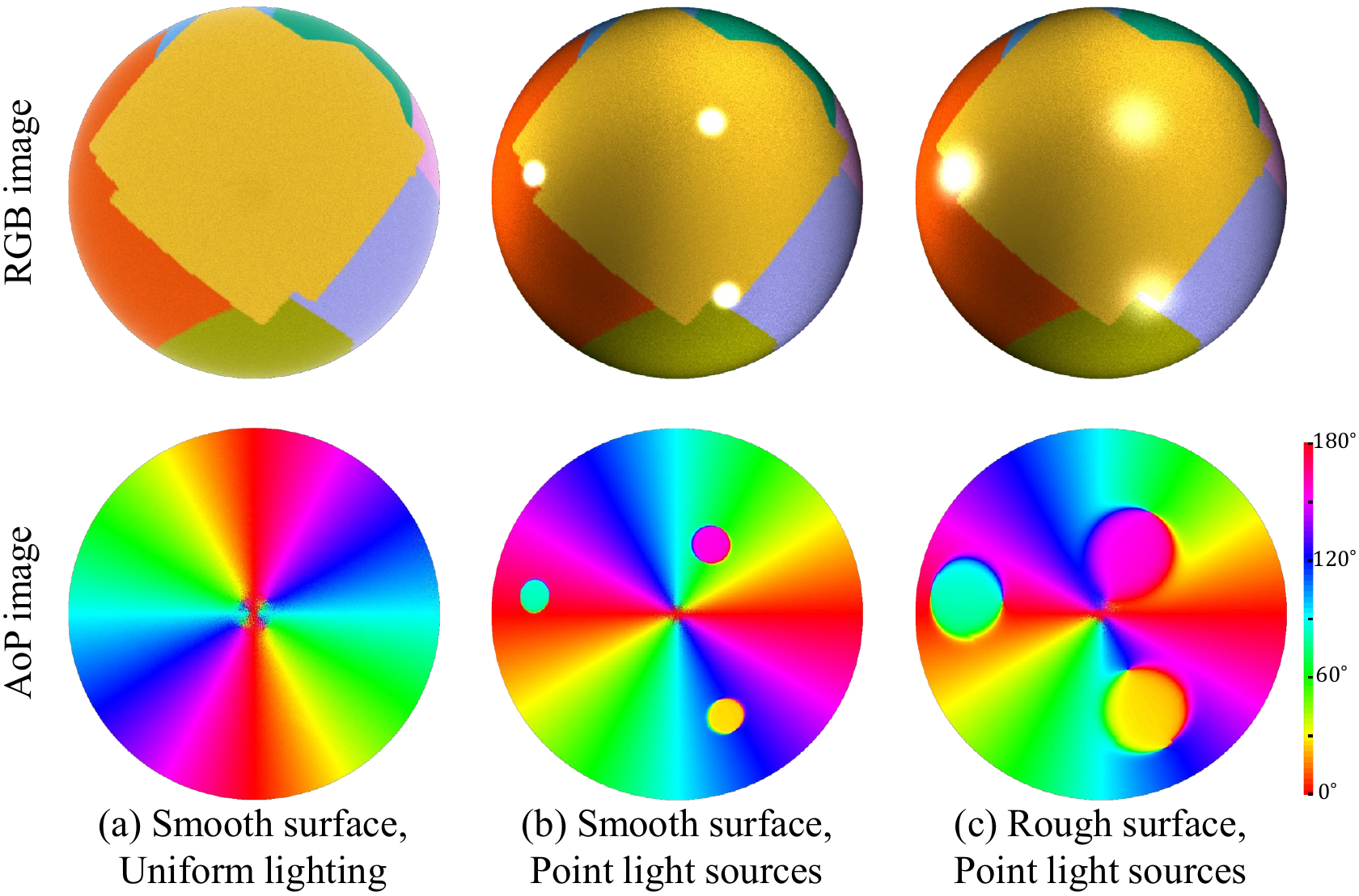}
   \caption{Examples of ambiguities in mixed polarization for a sphere shape: (a) Smooth surface under uniform lighting; (b) Smooth surface under point light sources; (c) Rough surface under point light sources. We can see that AoP values may differ even for the same point on the sphere depending on the lighting conditions and the surface material properties, meaning that there exist ambiguities between the AoP and the azimuth angle of the surface normal.
   }
   \label{fig:mixed}
\end{figure}

\begin{figure*}[b]
\centering
\includegraphics[trim={0cm 0cm 0cm 0cm}, width=1\linewidth]{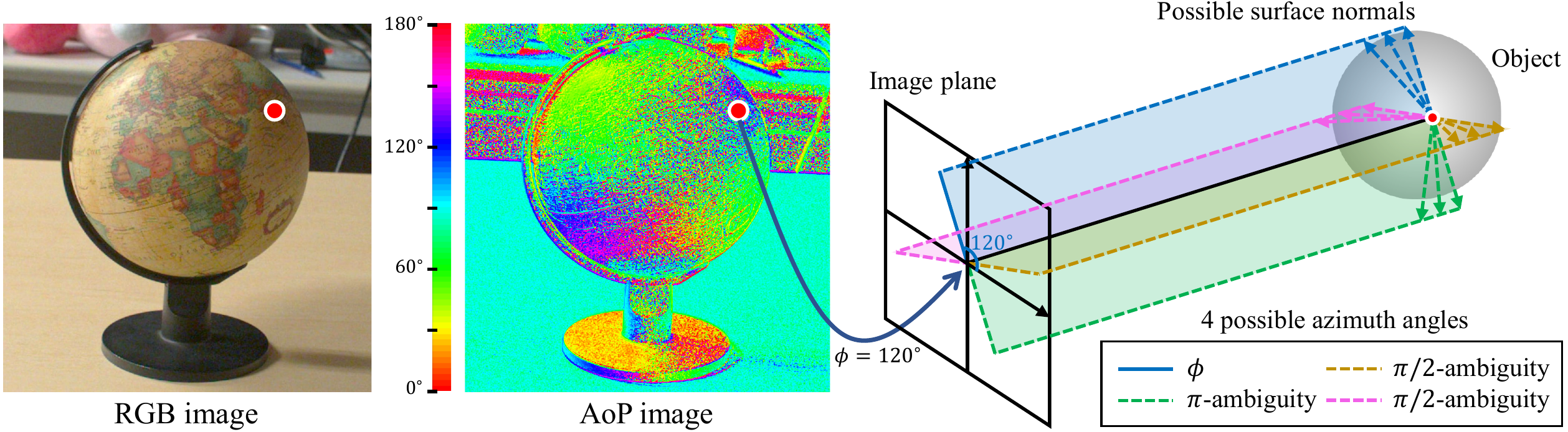}
\caption{Four possible azimuth angles ($\alpha=30^{\circ}$, $120^{\circ}$, $210^{\circ}$, and $300^{\circ}$) corresponding to an observed AoP value ($\phi=120^{\circ}$). The transparent color planes show the possible planes on which the surface normal has to lie. Example possible surface normals are illustrated by the color dashed arrows on the object.}
\label{fig:ambiguity}
\end{figure*}

Figure~\ref{fig:mixed} exemplifies this ambiguous relationship by using a sphere shape. 
Figure~\ref{fig:mixed}(a) is the observation of a smooth surface under uniform lighting, while Figs.~\ref{fig:mixed}(b) and~\ref{fig:mixed}(c) are the observations of a smooth surface and a rough surface under three point light sources, respectively.
Figure~\ref{fig:mixed}(a) shows the case that polarized specular reflection dominates under uniform lighting, while Figs.~\ref{fig:mixed}(b) and~\ref{fig:mixed}(c) show the cases that polarized diffuse reflection dominates in most parts under point light sources, except for the areas where strong specular reflections exist. 
Figures~\ref{fig:mixed}(a) and~\ref{fig:mixed}(b) show that the observed AoP information of a certain point may vary by $\pi/2$ when the lighting condition changes.
Figures~\ref{fig:mixed}(b) and~\ref{fig:mixed}(c) demonstrate that when the reflection property of the surface changes, the observed AoP information also may change. 
In short, because the observation of a certain point may change according to lighting conditions and an object's surface properties, there exist two kinds of ambiguities when estimating the surface normal.

\bm{$\pi$}{\bf -ambiguity:}
$\pi$-ambiguity exists because the range of AoP is from $0$ to $\pi$
while that of the azimuth angle is from $0$ to $2\pi$.
AoP corresponds to the same direction
or the inverse direction of the azimuth angle of the surface normal, i.e. AoP may be equal to the azimuth angle or have $\pi$'s difference with the azimuth angle.

\bm{$\pi/2$}{\bf -ambiguity:}
It is difficult to decide whether polarized specular reflection
or polarized diffuse reflection dominates without
any prerequisites for surface materials and lighting conditions.
AoP has $\pi/2$'s difference with the azimuth angle when polarized specular reflection dominates,
while it equals the azimuth angle or has $\pi$'s difference with the azimuth angle when polarized diffuse reflection dominates.
Therefore, there exists $\pi/2$-ambiguity in addition to $\pi$-ambiguity
when determining the relationship between the AoP and the azimuth angle.
The $\pi/2$-ambiguity becomes hard to hold on a rough surface, i.e. AoP becomes inconsistent with the angle that has $\pi/2$'s difference with the azimuth angle, as shown in Fig.~\ref{fig:mixed}(c). This is because AoP depends on the halfway vector between the viewing direction and the light direction, which is no longer close to the surface normal on a rough surface due to scattered microfacets. Therefore, our main focus in this paper is an object with a smooth surface.

Based on the above knowledge, as shown in Fig.~\ref{fig:ambiguity}, for the AoP value ($\phi=120^\circ$) of the pixel marked in red, four possible azimuth angles (i.e. $\alpha=30^{\circ}$, $120^{\circ}$, $210^{\circ}$, and $300^{\circ}$) can be inferred as depicted by the four lines on the image plane.
The planes where the surface normal has to lie, which are represented by the four transparent color planes, are determined according to the four possible azimuth angles. The dashed arrows on the object show the examples of possible surface normals, which are constrained on the planes. In our method, the explained relationship between the AoP measurement and the possible azimuth angles is exploited to constrain the estimated surface vertex's normal.

\begin{figure*}[t!]
   \centering
   \includegraphics[trim={0cm 0cm 0cm 0cm}, width=1\linewidth]{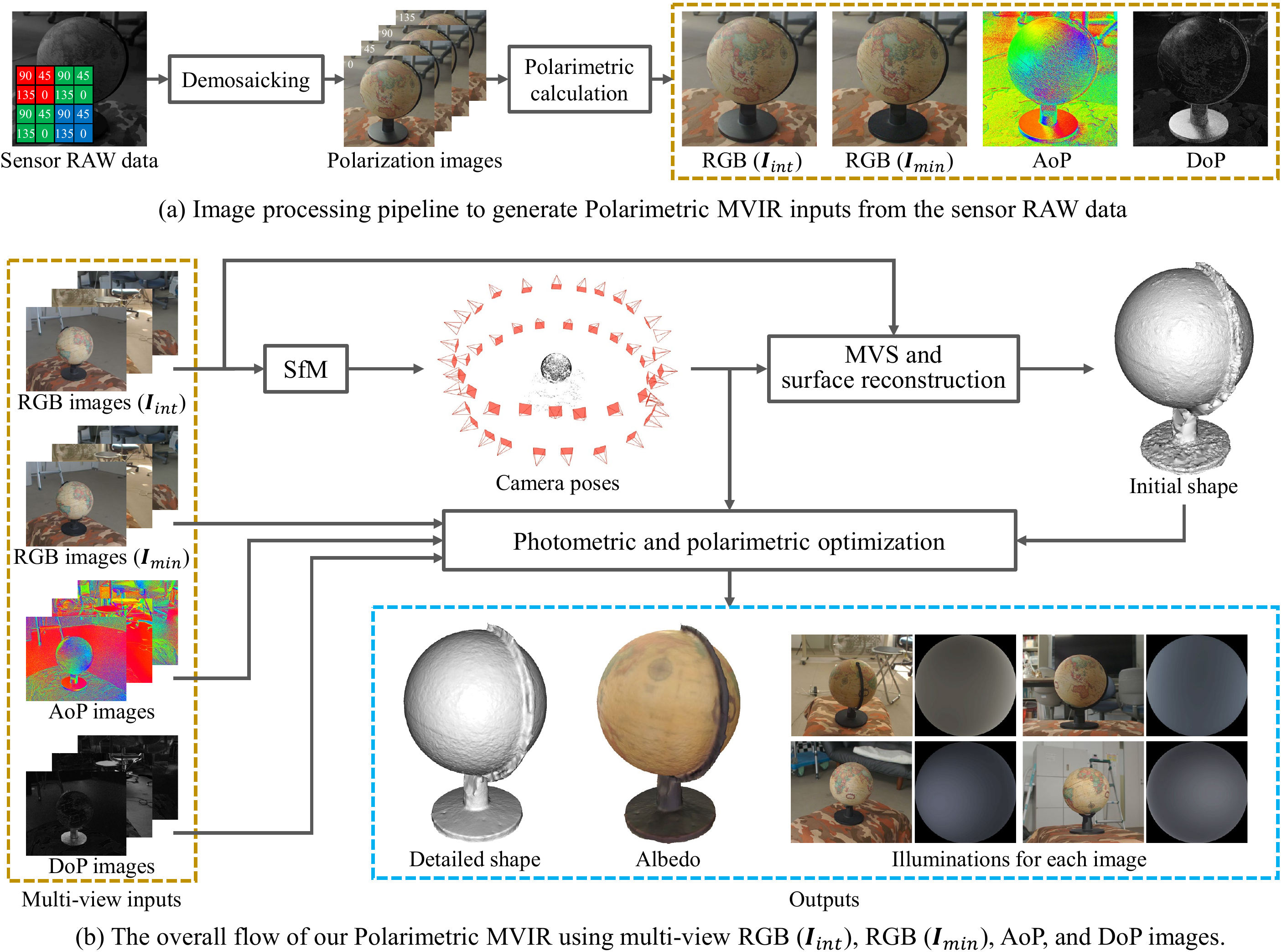}
   \caption{The overall flows of (a) the image processing pipeline to generate Polarimetric MVIR inputs from the sensor raw data and (b) our Polarimetric MVIR using multi-view RGB (${\bm I}_{int}$), RGB (${\bm I}_{min}$), AoP, and DoP images.}
   \label{fig:flowchart}
   \label{flowchart-test}
\end{figure*}

\section{Polarimetric MVIR}
\label{sec:pmvir}

\subsection{Color-polarization sensor data processing}
\label{subsec:rawDataProcessing}

Figure~\ref{fig:flowchart}(a) shows the flowchart to obtain input RGB, AoP, and DoP images from sensor raw data, where we use a one-shot color-polarization camera consisting of a $4\!\times\!4$ regular pixel pattern~\cite{maruyama20183}, though our method is not limited to this kind of polarization camera. For every pixel, twelve values, which are four sets of RGB values, i.e. $(R,G,B)\times (I_0,I_{45},I_{90},I_{135})$, denoted as $\left({\bm I}_{0}, {\bm I}_{45}, {\bm I}_{90}, {\bm I}_{135}\right)$ are obtained by interpolating the raw mosaic data.
As proposed in \cite{morimatsu2021monochrome}, the pixel values for each polarization direction in every $2\!\times\!2$ block are extracted to obtain Bayer-patterned data for that direction. Then, Bayer color interpolation~\cite{kiku2016beyond} and polarization interpolation~\cite{mihoubi2018survey} are sequentially performed to obtain 12-channel full-color-polarization data.

We then calculate the RGB intensity ${\bm I}_{int}=\left[I_{R},I_{G},I_{B}\right]^T$ as the average of four polarization directions as 
\begin{equation}
\label{eq:I}
{\bm I}_{int} = ({\bm I}_{0}+{\bm I}_{45}+{\bm I}_{90}+{\bm I}_{135})/4.
\end{equation}
AoP and DoP values are also calculated according to Eq.(\ref{eq:Stokes}), (\ref{eq:AoP}), and (\ref{eq:DoP}), using the average values of $\left(\bar{I}_0, \bar{I}_{45}, \bar{I}_{90}, \bar{I}_{135}\right)$, which represent the average of RGB values for each polarization direction. 

Since the unpolarized part of the reflection can suppress the influence of specular reflection~\cite{atkinson2006recovery}, it is beneficial for our photometric optimization.
Therefore, for the RGB images used for the optimization, we employ unpolarized RGB values ${\bm I}_{min}=\left[I_{R_{min}},I_{G_{min}},I_{B_{min}}\right]^T$, whose elements correspond to ${I}_{min}$ in R, G, and B channels, respectively. ${\bm I}_{min}$ can be obtained as 
\begin{equation}
\label{eq:Imin}
   {\bm I}_{min} = (1-\rho) \cdot {\bm I}_{int},
\end{equation}
where $\rho$ is DoP. The examples of the derived RGB~(${\bm I}_{int}$), RGB~(${\bm I}_{min}$), AoP, and DoP images are shown in Fig.~\ref{fig:flowchart}(a).

\subsection{Initial geometric reconstruction}
Figure~\ref{fig:flowchart}(b) shows the overall flow of our Polarimetric MVIR using multi-view RGB (${\bm I}_{int}$), RGB (${\bm I}_{min}$), AoP, and DoP images. It starts with initial geometric 3D reconstruction as follows. SfM is firstly performed using the standard RGB  (${\bm I}_{int}$) images to estimate camera poses. Then, MVS and surface reconstruction are applied to obtain an initial surface model, which is represented by a triangular mesh. The visibility of each vertex to each camera, which is used to select the cameras applied for the optimization to each vertex, is then checked using the algorithm in~\cite{kim2016multi}. Finally, to increase the number of vertices, the initial surface is subdivided by $\sqrt{3}$-subdivision~\cite{kobbelt20003} until the maximum pixel area in each triangular patch projected to visible cameras becomes smaller than a threshold. The visibility of newly generated vertices also can be determined according to the visibility of each mesh before the subdivision.

\subsection{Photometric and polarimetric optimization}
\label{sec:optimize}
The photometric and polarimetric optimization is then performed to refine the initial model while estimating each vertex's albedo and each image's illumination. The cost function is expressed as
\begin{equation}
   \label{eq:costFunction}
   \begin{aligned}
      \mathop{\arg\min}\limits_{{\bf X}, {\bf K}, {\bf L}} \; & E_{pho}({\bf X}, {\bf K}, {\bf L}) + \tau_1 E_{pol}({\bf X}) \\
      & + \tau_2 E_{gsm}({\bf X}) + \tau_3 E_{psm}({\bf X,\bf K}),
   \end{aligned}
\end{equation}
where $E_{pho}$, $E_{pol}$, $E_{gsm}$, and $E_{psm}$ represent a photometric rendering term, a polarimetric term, a geometric smoothness term, and a photometric smoothness term, respectively. $\tau_1$, $\tau_2$, and $\tau_3$ are weights to balance each term. Similar to MVIR~\cite{kim2016multi}, the optimization parameters are defined as below:

\noindent
- ${\bf X}\in\mathbb{R}^{3\times m}$ is the
vertex 3D coordinate set, where $m$ is the total number
of vertices.

\noindent
- ${\bf K}\in\mathbb{R}^{3\times m}$ is the vertex albedo set, where each vertex albedo is expressed in the RGB color space as ${\bm K} =[K_R,K_G,K_B]^T$.

\noindent
- ${\bf L}\in\mathbb{R}^{12\times n}$ is the scene illumination matrix, where $n$
is the total number of images. Each image's illumination parameters are represented as ${\bm L} = [{\bm L}_{basis}; {\bm L}_{scale}]$ by using nine coefficients for the second-order spherical harmonics basis ${\bm L}_{basis} = [L_0,\cdots,L_8]^T$ and three RGB color scales ${\bm L}_{scale} =[L_R,L_G,L_B]^T$.

\subsubsection{Photometric rendering term}
We adopt the same photometric rendering term as MVIR, which is expressed as
\begin{equation}
   \label{photometricRenderingTerm}
   E_{pho}({\bf X}, {\bf K}, {\bf L})=\sum_i \sum_{c\in \mathcal{V}(i)} \frac{
   ||{\bm I}_{i,c}({\bf X})-\hat{{\bm I}}_{i,c}({\bf X}, {\bf K}, {\bf L})||^2}{|\mathcal{V}(i)|},
\end{equation}
which measures the pixel-wise difference between observed and rendered RGB values.
${\bm I}_{i,c}\in\mathbb{R}^3$ is the observed RGB values, for which we use the RGB~(${\bm I}_{min})$ image, of the pixel in $c$-th image corresponding to $i$-th vertex's projection and $\hat{{\bm I}}_{i,c}\in \mathbb{R}^3$ is the corresponding rendered RGB values.
$\mathcal{V}(i)$ represents the visible camera set for $i$-th vertex. The perspective projection model is used to project each vertex to each camera.
Suppose ${\bm K} =[K_R,K_G,K_B]^T$ represent the albedo for $i$-th vertex and ${\bm L}_{basis} = [L_0,\cdots,L_8]^T$ and ${\bm L}_{scale} =[L_R,L_G,L_B]^T$ represent the illumination parameters for $c$-th image, where the indexes $i$ and $c$ are omitted for notation simplicity.
The rendered RGB values are then calculated as
\begin{equation}
   \label{eq:rendering}
   \hat{{\bm I}}_{i,c}({\bf X}, {\bf K}, {\bf L})=
   \left[\begin{matrix}
         K_RS({\bm N}({\bf X}),{\bm L}_{basis})L_R \\
         K_GS({\bm N}({\bf X}),{\bm L}_{basis})L_G \\
         K_BS({\bm N}({\bf X}),{\bm L}_{basis})L_B
      \end{matrix}
      \right],
\end{equation}
where $S$ is the shading calculated by using the second-order spherical harmonics illumination model~\cite{ramamoorthi2001efficient,wu2011high} as
\begin{equation}
   \label{eq:shading}
   \begin{aligned}
      S({\bm N}({\bf X}),{\bm L}_{basis})= & \;L_0+L_1N_y+L_2N_z+L_3N_x                  \\
                             & +L_4N_xN_y+L_5N_yN_z+L_6(N_z^2-\frac{1}{3}) \\
                             & +L_7N_xN_z+L_8(N_x^2-N_y^2),
   \end{aligned}
\end{equation}
where ${\bm N}({\bf X}) = [N_x,N_y,N_z]^T$ represents the vertex's normal vector, which is calculated as the average of adjacent triangular patch's normals. By this model, varying illuminations for each image and spatially-varying albedos are considered.

\begin{figure}[t!]
   \centering
   \includegraphics[trim={0cm 0cm 0cm 0cm}, width=1\linewidth]{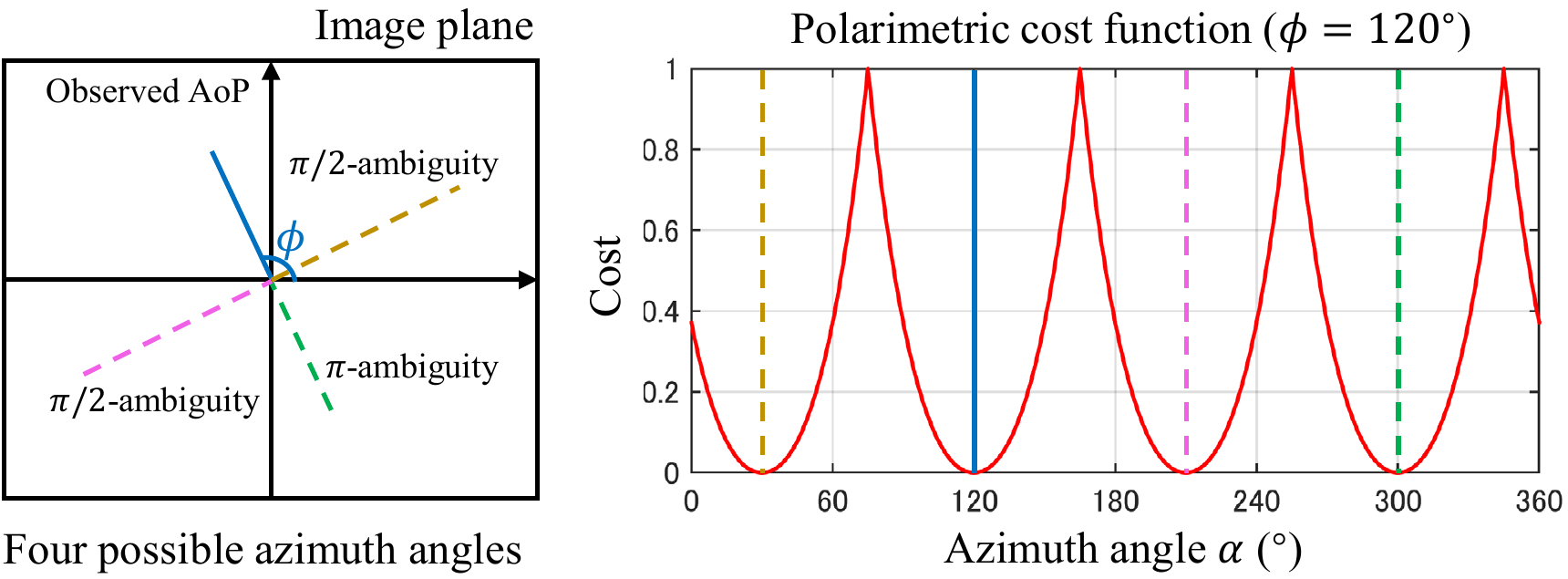}
   \caption{An example of the polarimetric cost function
   ($\phi=120^{\circ}, k=0.5$). Four lines correspond to
   possible azimuth angles as shown in Fig.~\ref{fig:ambiguity}.}
   \label{fig:cost}
\end{figure}

\subsubsection{Polarimetric term} To effectively constrain each estimated surface vertex's normal, we here propose a novel polarimetric term. Figure~\ref{fig:cost} shows an example of our polarimetric cost function for the case that the AoP measurement of the pixel corresponding to the vertex’s projection equals 120$^\circ$, i.e. $\phi=120^\circ$. This example corresponds to the situation as shown in Fig.~\ref{fig:ambiguity}. In both figures, four possible azimuth angles derived from the AoP measurement are shown by blue solid, purple dashed, green dashed, and brown dashed lines on the image plane, respectively. These four possibilities are caused by both the $\pi$-ambiguity and the $\pi/2$-ambiguity introduced in Section~\ref{sec:ambiguities}. In the ideal case without noise, one of the four possible azimuth angles should be the same as the azimuth angle of (unknown) true surface normal.

Based on this principle, as shown in Fig.~\ref{fig:cost}, our polarimetric term evaluates the difference between the azimuth angle of the estimated surface vertex's normal $\alpha$ and its closest possible azimuth angle from the AoP measurement~(i.e. $\phi-\pi/2,\ \phi,\ \phi+\pi/2$, or $\phi+\pi$). The cost function is mathematically defined as
   \begin{equation}
      \label{eq:pol}
      E_{pol}({\bf X})=\sum_i \sum_{c\in \mathcal{V}(i)}
      \rho_{i,c}({\bf X})\left(\frac{e^{-k\theta_{i,c}({\bf X})}-e^{-k}}{1-e^{-k}}\right)^2/{|\mathcal{V}(i)|},
   \end{equation}
where $k$ is a parameter that determines the narrowness of the concave to assign the cost (see the example in Fig.~\ref{fig:cost}). $\rho_{i,c}$ is the DoP weighting, which assigns a higher weight for a larger DoP value, i.e. a higher reliability of the polarimetric information. Since the observed DoP values for a certain vertex vary according to viewing directions, the DoP weighting can also be served to select the viewing directions reliable to constrain the vertex normal.
$\theta_{i,c}$ is defined as
\begin{equation}
\label{eq:theta}
\begin{aligned}
\theta_{i,c}({\bf X})=1-4\eta_{i,c}({\bf X})/\pi,
\end{aligned}
\end{equation}
where $\eta_{i,c}$ is expressed as
\begin{equation}
\label{eq:eta}
\begin{aligned}
\eta_{i,c}({\bf X})=\mathop{\min}( & |\alpha_{i,c}({\bm N(\bf X)})-\phi_{i,c}({\bf X})-2\pi|, \\
& |\alpha_{i,c}({\bm N(\bf X)})-\phi_{i,c}({\bf X})-3\pi/2|,       \\
& |\alpha_{i,c}({\bm N(\bf X)})-\phi_{i,c}({\bf X})-\pi|,       \\
& |\alpha_{i,c}({\bm N(\bf X)})-\phi_{i,c}({\bf X})-\pi/2|,       \\
& |\alpha_{i,c}({\bm N(\bf X)})-\phi_{i,c}({\bf X})|,       \\
& |\alpha_{i,c}({\bm N(\bf X)})-\phi_{i,c}({\bf X})+\pi/2|, \\
& |\alpha_{i,c}({\bm N(\bf X)})-\phi_{i,c}({\bf X})+\pi|).
\end{aligned}
\end{equation}
Here, $\alpha_{i,c}$ is the azimuth angle calculated by the projection of $i$-th vertex's normal to $c$-th image plane and $\phi_{i,c}$ is the corresponding AoP measurement.

Our polarimetric term mainly has two benefits. First, it enables us to constrain the estimated surface vertex's normal while simultaneously resolving the ambiguities based on the optimization using all vertices and all multi-view AoP and DoP measurements. Second, the concave shape of the cost function makes the normal constraint more robust to noise, which is an important property since AoP is susceptible to noise. The balance between the strength of the normal constraint and the robustness to noise can be adjusted by the parameter~$k$.

\subsubsection{Geometric smoothness term}
The geometric smoothness term is applied to regularize the cost and to derive a smooth surface. This term is described as
\begin{equation}
   \label{eq:gsm}
   E_{gsm}({\bf X})=\sum_{r}
   \left(\frac{{\rm arccos}\left({\bm N}^\prime_r({\bf X})\cdot {\bm N}^\prime_{r_{avg}}({\bf X})\right)}{\pi}\right)^{t},
\end{equation}
where ${\bm N}^\prime_r$ represents the normal of $r$-th triangular patch, ${\bm N}^\prime_{r_{avg}}$ represents the averaged normal of its adjacent patches, and $t$ is a parameter to assign the cost. This term becomes small if the curvature of the surface is close to constant.

\subsubsection{Photometric smoothness term}
Changes of pixel values in each image may result from different albedos or shading since spatially varying albedos are allowed in our model. To regularize this uncertainty, the same photometric smoothness term as \cite{kim2016multi} is applied as
\begin{equation}
   E_{psm}({\bf X,\bf K})=\sum_{i} \sum_{j\in \mathcal{A}(i)}
   w_{i,j}({\bf X})\left|\left|({\bm K}_i-{\bm K}_j)\right|\right|^2,
\end{equation}
where $\mathcal{A}(i)$ is the set of adjacent vertices of $i$-th vertex and $w_{i,j}$ is the weight for the pair of $i$-th and $j$-th vertices.
   A small weight $w_{i,j}$ is assigned,
i.e. change of albedo is allowed, if a large chromaticity or intensity difference is observed between the corresponding pixels in the RGB image (see~\cite{kim2016multi} for details). By this term, a smooth variation in photometric information is considered as the result of shading while a sharp variation is considered as the result of varying albedos.

\section{Experimental Results}
\subsection{Implementation details}
\label{subsec:details}
We apply COLMAP~\cite{schonberger2016structure} for SfM to estimate camera poses and OpenMVS~\cite{OpenMVS} for MVS to obtain a dense point cloud. The initial surface for our method is reconstructed from the point cloud by the built-in surface reconstruction function of OpenMVS. 
The albedo of each vertex is initialized as the average of the pixel values of all visible cameras. The illumination ${\bm L} = [{\bm L}_{basis}; {\bm L}_{scale}]$ is initialized as ${\bm L}_{basis} = [1,0,0,\cdots,0]^T$, which represents a uniform environment map, and ${\bm L}_{scale} =[1,1,1]^T$, which represents a white illumination.
Given the camera poses and the above initial values, the cost optimization of Eq.~(\ref{eq:costFunction}) is iterated three times by changing the weight parameters as $(\tau_1, \tau_2, \tau_3)$ = $(60.0, 0.1, 2.0)$, $(120.0, 0.1, 2.0)$, and $(360.0, 0.1, 2.0)$. For each iteration, the parameter $t$ in Eq.~(\ref{eq:gsm}) is changed as $t$ = $2.2$, $2.8$, and $3.4$, while the parameter
$k$ in Eq.~(\ref{eq:pol}) is set to 0.5 in all the iterations.

By these iterations, the surface normal constraint from AoP is gradually strengthened, while the geometric smoothness constraint is gradually weakened. This allows subtle variations between the normals of adjacent vertices to derive a fine shape while avoiding a local minimum. The non-linear optimization problem is solved by using DynamicNumericDiffCostFunction in Ceres solver~\cite{ceres-solver} to derive the numerical differentiation, where the 3D coordinates and the albedos of all vertices, and the scene illumination matrix are optimized to minimize the cost function.

\subsection{Comparison using synthetic data}
\setlength{\tabcolsep}{4pt}
\begin{table*}[t!]
\begin{center}
   \renewcommand\arraystretch{1.2}
\setlength\aboverulesep{0pt}
\setlength\belowrulesep{0pt}
\setcellgapes{1.5pt}\makegapedcells
\caption{Comparisons of the average accuracy (Acc.) and completeness (Comp.) errors}
\label{table:evaluation}
\begin{tabular}{c|p{7em}|l|l|l|l|l|l}
   \toprule
   \multicolumn{1}{c}{} & \multicolumn{1}{c|}{} & \multicolumn{1}{p{6em}|}{PMVS} & \multicolumn{1}{p{6em}|}{CMPMVS} & \multicolumn{1}{p{6em}|}{COLMAP} & \multicolumn{1}{p{6em}|}{OpenMVS} & \multicolumn{1}{p{6em}|}{MVIR} & \multicolumn{1}{p{12em}}{Polarimetric MVIR (Ours)} \\
   \midrule
   \midrule
   \multirow{3}[6]{*}{Armadillo} & \multicolumn{1}{l|}{\# of Vertices} & 52,615  & 198,391  & 286,278  & 1,752,640  & 303,139  & 303,137  \\
   \cmidrule{2-8}      & Acc.($\times10^{-2}$) & 1.038  & 0.815  & 0.745  & 0.918  & 0.840  & \textcolor{red}{\textbf{0.395 }} \\
   \cmidrule{2-8}      & Comp.($\times10^{-2}$) & 2.068  & 1.440  & 1.345  & 0.625  & 1.005  & \textcolor{red}{\textbf{0.599 }} \\
   \midrule
   \multirow{3}[6]{*}{Bunny} & \multicolumn{1}{l|}{\# of Vertices} & 59,508  & 88,115  & 186,396  & 1,526,552  & 299,629  & 299,628  \\
   \cmidrule{2-8}      & Acc.($\times10^{-2}$) & 1.319  & 1.148  & \textcolor{red}{\textbf{0.829 }} & 1.120  & 1.646  & 0.880  \\
   \cmidrule{2-8}      & Comp.($\times10^{-2}$) & 4.151  & 3.779  & 5.352  & 1.462  & 1.876  & \textcolor{red}{\textbf{1.126 }} \\
   \midrule
   \multirow{3}[6]{*}{Dragon} & \multicolumn{1}{l|}{\# of Vertices} & 64,107  & 145,727  & 282,925  & 1,711,714  & 363,152  & 363,117  \\
   \cmidrule{2-8}      & Acc.($\times10^{-2}$) & 1.356  & 1.417  & 0.902  & 1.133  & 1.274  & \textcolor{red}{\textbf{0.738 }} \\
   \cmidrule{2-8}      & Comp.($\times10^{-2}$) & 3.529  & 3.717  & 4.053  & 1.954  & 2.072  & \textcolor{red}{\textbf{1.561 }} \\
   \midrule
   \midrule
   \multirow{2}[4]{*}{Average} & Acc.($\times10^{-2}$) & 1.238  & 1.127  & 0.825  & 1.057  & 1.253  & \textcolor{red}{\textbf{0.671 }} \\
   \cmidrule{2-8}      & Comp.($\times10^{-2}$) & 3.250  & 2.979  & 3.583  & 1.347  & 1.651  & \textcolor{red}{\textbf{1.095 }} \\
   \bottomrule
   \end{tabular}%   
\end{center}  
\end{table*}%
\setlength{\tabcolsep}{1.4pt}
 
\subsubsection{Synthetic data generation}
Numerical evaluation was performed using three CG models (Armadillo, Stanford bunny, and Dragon) available from Stanford 3D Scanning Repository~\cite{Stanford}.
The original 3D models were subdivided to provide a sufficient number of vertices as ground truths. 
We synthesized RGB, AoP, and DoP inputs using Mitsuba 2 renderer~\cite{nimier2019mitsuba}, which supports a pBRDF reflection model~\cite{baek2018simultaneous} to simulate realistic polarization images. 
In Mistuba 2 renderer, the polarization rendering can be applied to some specific materials, such as a dielectric material and conductor.
Among these materials, we used a polarized plastic material for the comparisons in Section~\ref{sec:compare} and \ref{sec:ablation} because it supports spatially-varying vertex albedos. The roughness parameter of the plastic material was set to $0.0$, which represents the smoothest surface and exhibits relatively high DoPs.
In addition to the plastic material, we also used the unpolarized Lambert material and the gold material, which is a common type of conductor, in Section~\ref{sec:material} to evaluate the robustness of our method to different levels of DoPs and ambiguities that depend on the materials. 

\begin{figure}[t!]
   \centering
   \includegraphics[trim={0cm 0cm 0cm 0cm}, width=1.0\linewidth]{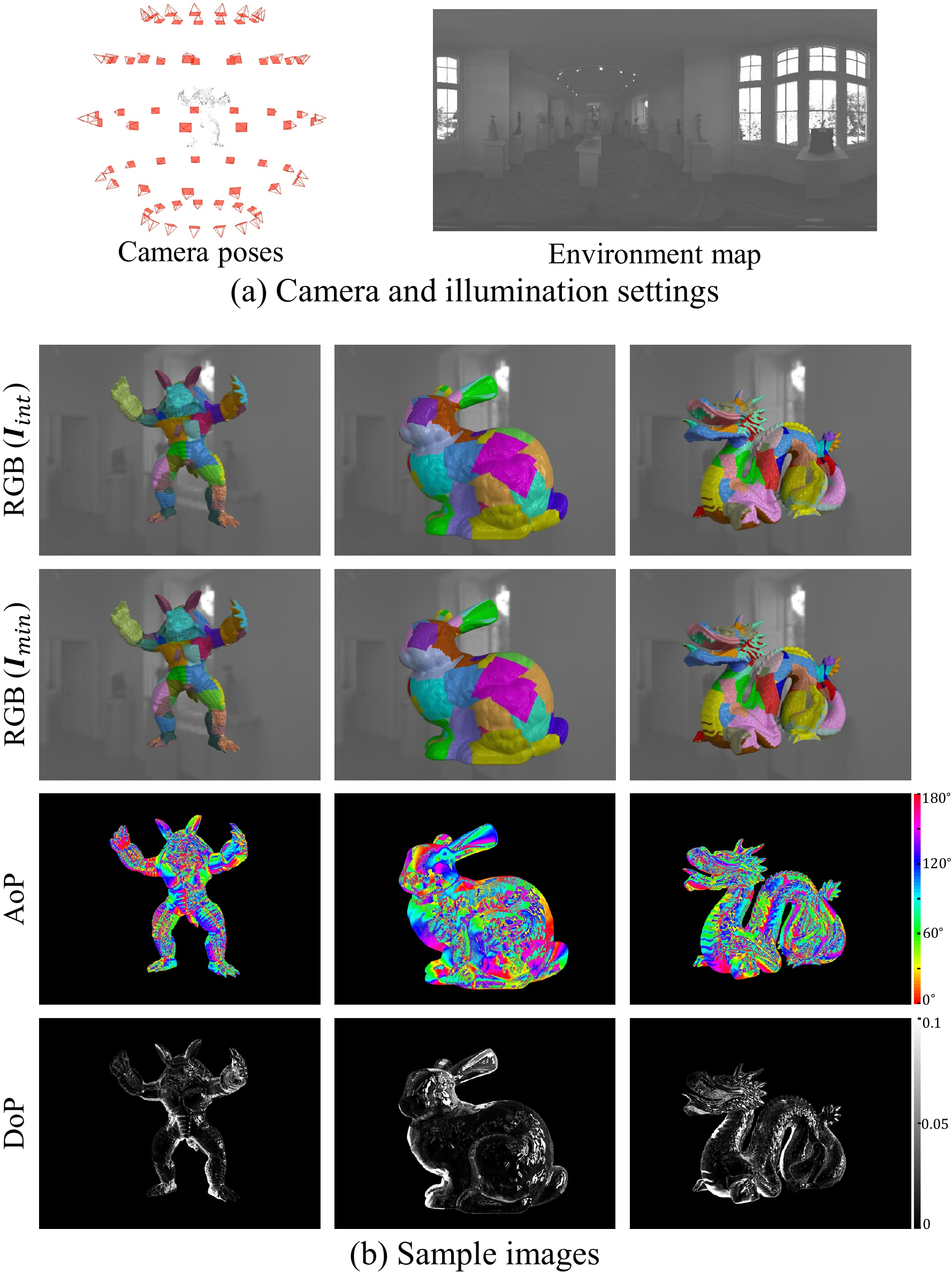}\\
   \vspace{-1mm}
   \caption{Synthetic data generation. (a) Camera and illumination settings and (b) the examples of the synthesized images using the plastic material.}
   \label{fig:setting}
\end{figure}

To render polarization images, we used spherically-placed cameras and an environment map in~\cite{envmap}, as shown in Fig.~\ref{fig:setting}(a). For each camera view, a polarizer was placed in front of the camera and rotated to obtain the polarization images of the directions of $0^\circ$, $45^\circ$, $90^\circ$, and $135^\circ$. Using those images, RGB~(${\bm I}_{int}$), RGB~(${\bm I}_{min}$), AoP, and DoP images were obtained according to the calculations in Section~\ref{subsec:rawDataProcessing}. The examples of the generated images using the plastic material are shown in Fig.~\ref{fig:setting}(b), where we applied synthetic textures for the numerical evaluation of albedos.

\subsubsection{Comparison with existing methods}
\label{sec:compare}
We compared our Polarimetric MVIR with four representative MVS methods (PMVS~\cite{furukawa2009accurate}, CMPMVS~\cite{jancosek2011multi}, MVS in COLMAP~\cite{schonberger2016pixelwise}, and OpenMVS~\cite{OpenMVS}) and MVIR~\cite{kim2016multi}. For this comparison, ground-truth camera poses were used to avoid the alignment problem among the resultant models reconstructed from each method. MVIR and our method applied the same initial model, which was reconstructed by OpenMVS. To assess the shape quality, two commonly-used metrics~\cite{aanaes2016large,ley2016syb3r}, i.e. ``Accuracy" which is the distance from each estimated 3D point to its nearest ground-truth 3D point and ``Completeness" which is the distance from each ground-truth 3D point to its nearest estimated 3D point, were used. As the estimated 3D points, each 3D point of the output point cloud was evaluated for PMVS, COLMAP, and OpenMVS, while each vertex of the output surface was evaluated for CMPMVS, MVIR, and our method, according to each method's output form. 

Table~\ref{table:evaluation} shows the comparison of the average accuracy and the average completeness for each model. 
Among existing MVS methods, COLMAP achieves the best accuracy, but the worst completeness.
OpenMVS achieves the best completeness and it is also the best-balanced MVS method when considering both the accuracy and the completeness.
Comparing the results of OpenMVS and MVIR for Bunny and Dragon, MVIR fails to refine the initial shape~(OpenMVS's result) and shows lower accuracy and completeness results. In contrast, our method successfully refines the shape for all three models and achieves the best average accuracy and completeness with significant improvements, which demonstrates the effectiveness of our method under the existence of polarized reflections.
Regarding the comparison of MVIR and our method, we will provide more results and discussion in Section~\ref{sec:material}.

\begin{figure*}[t!]
   \centering
   \includegraphics[trim={0cm 0cm 0cm 0cm}, width=1.0\linewidth]{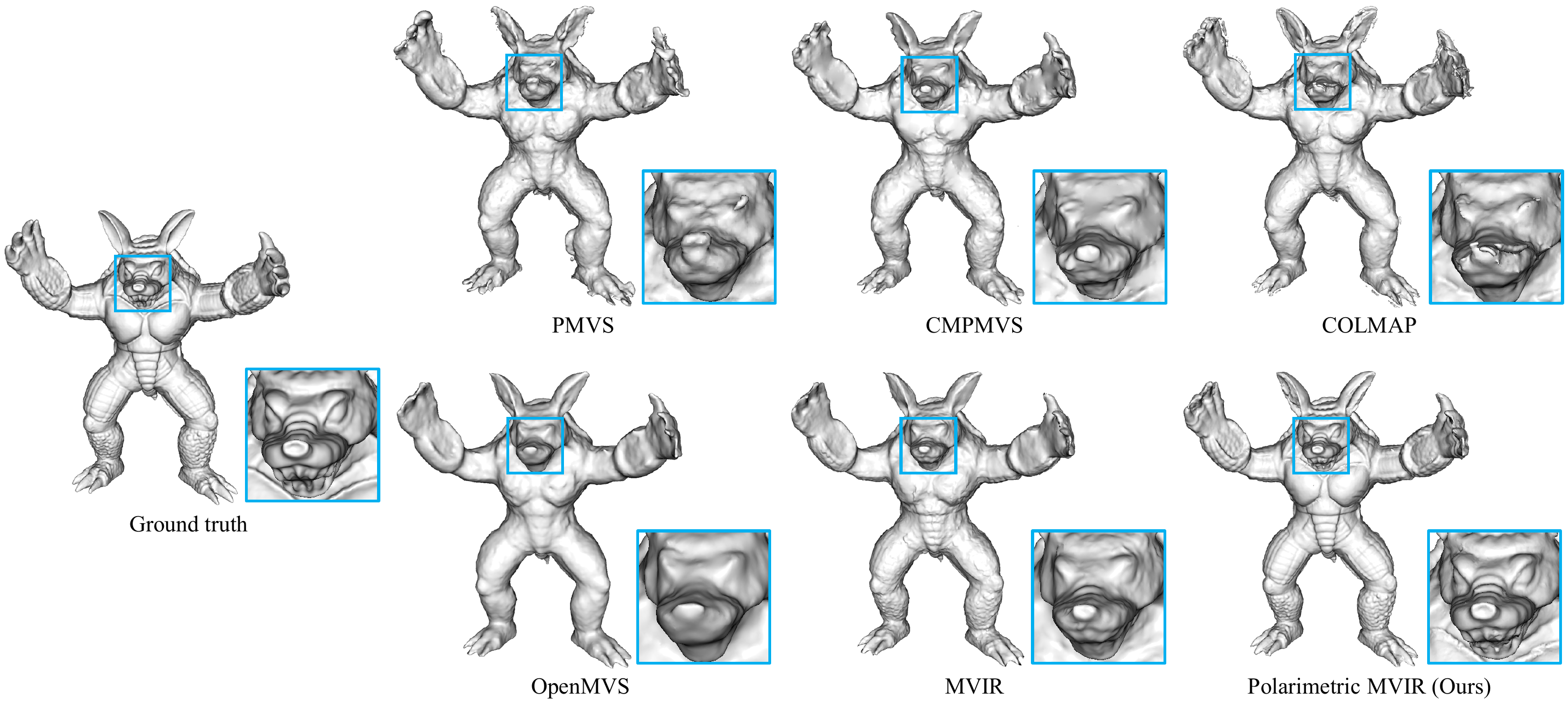}
   \caption{Visual comparison for the Armadillo model.}
   \label{fig:evaluation}
\end{figure*}

\begin{figure}[t!]
   \centering
   \includegraphics[trim={0cm 0cm 0cm 0cm}, width=1.0\linewidth]{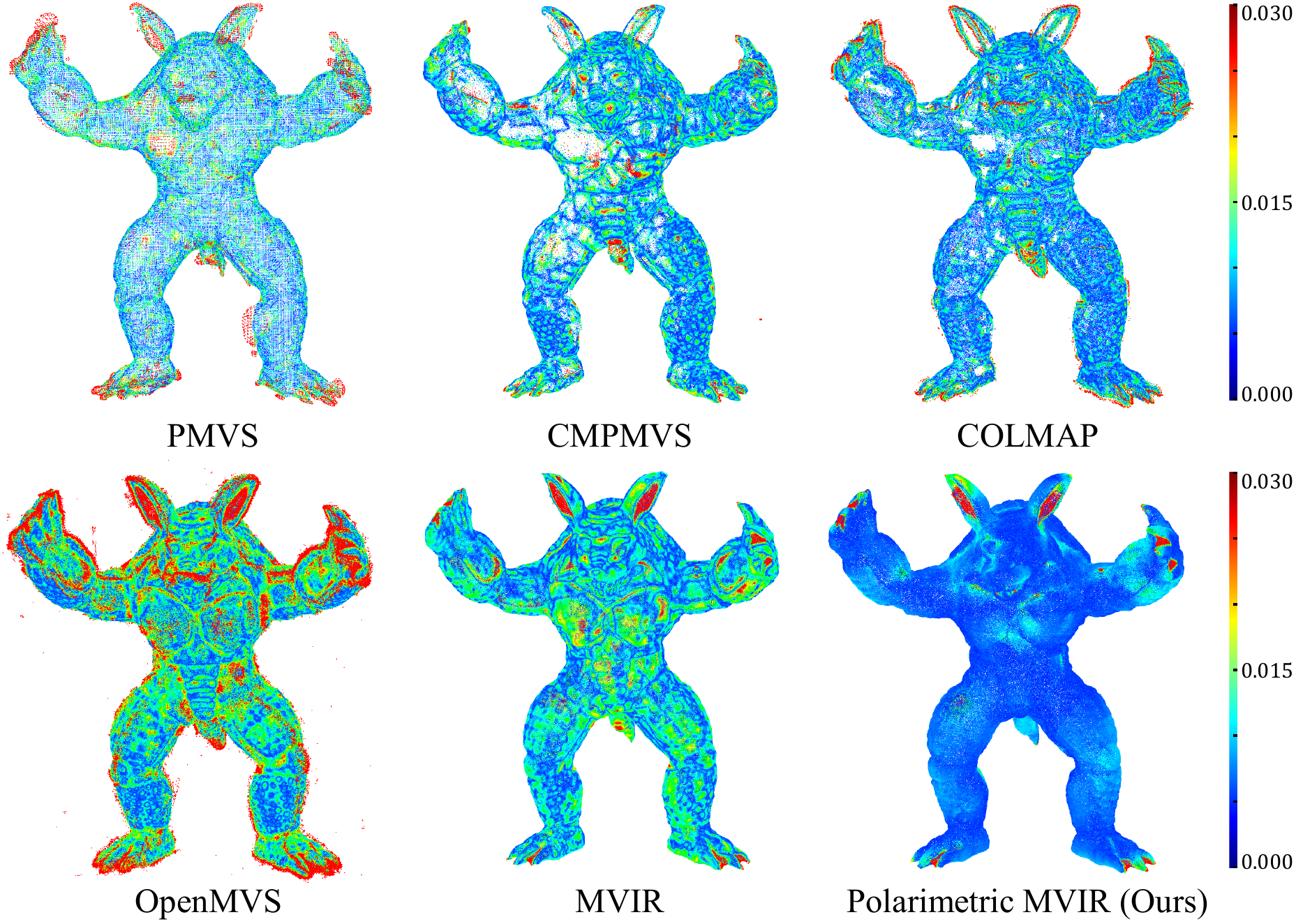}
   \caption{Accuracy error map for the Armadillo model.}
   \label{fig:accuracy}
\end{figure}

\begin{figure}[t!]
   \centering
   \includegraphics[trim={0cm 0cm 0cm 0cm}, width=1.0\linewidth]{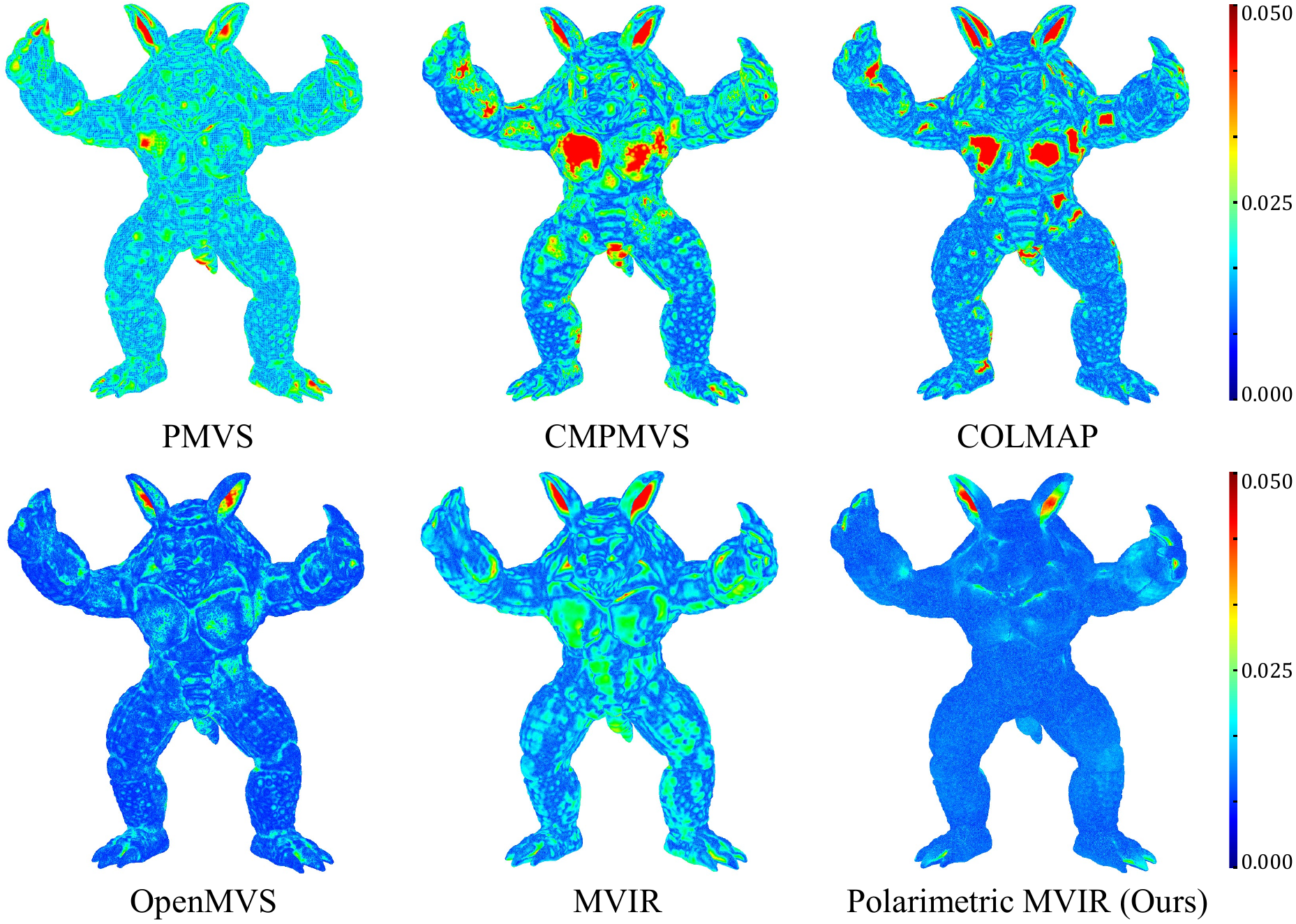}
   \caption{Completeness error map for the Armadillo model.}
   \label{fig:completeness}
\end{figure}

The visual comparison for Armadillo is shown in Fig.~\ref{fig:evaluation}, where the surfaces for PMVS and COLMAP were created using Screened Poisson Surface Reconstruction~\cite{kazhdan2013screened} with our best parameter choice, while the surface for OpenMVS was obtained using its built-in surface reconstruction function. We can clearly see that our method can recover more details than the other methods by exploiting polarimetric information, especially in the face part of the Armadillo model. For better visualization, the accuracy map and the completeness map are shown in Figs.~\ref{fig:accuracy} and~\ref{fig:completeness}, respectively, which illustrate that our method achieves the least errors.
The visual comparison for the Bunny and Dragon models can be seen in our supplementary material.

\begin{figure}[t!]
   \centering
   \includegraphics[trim={0cm 0cm 0cm 0cm}, width=1.0\linewidth]{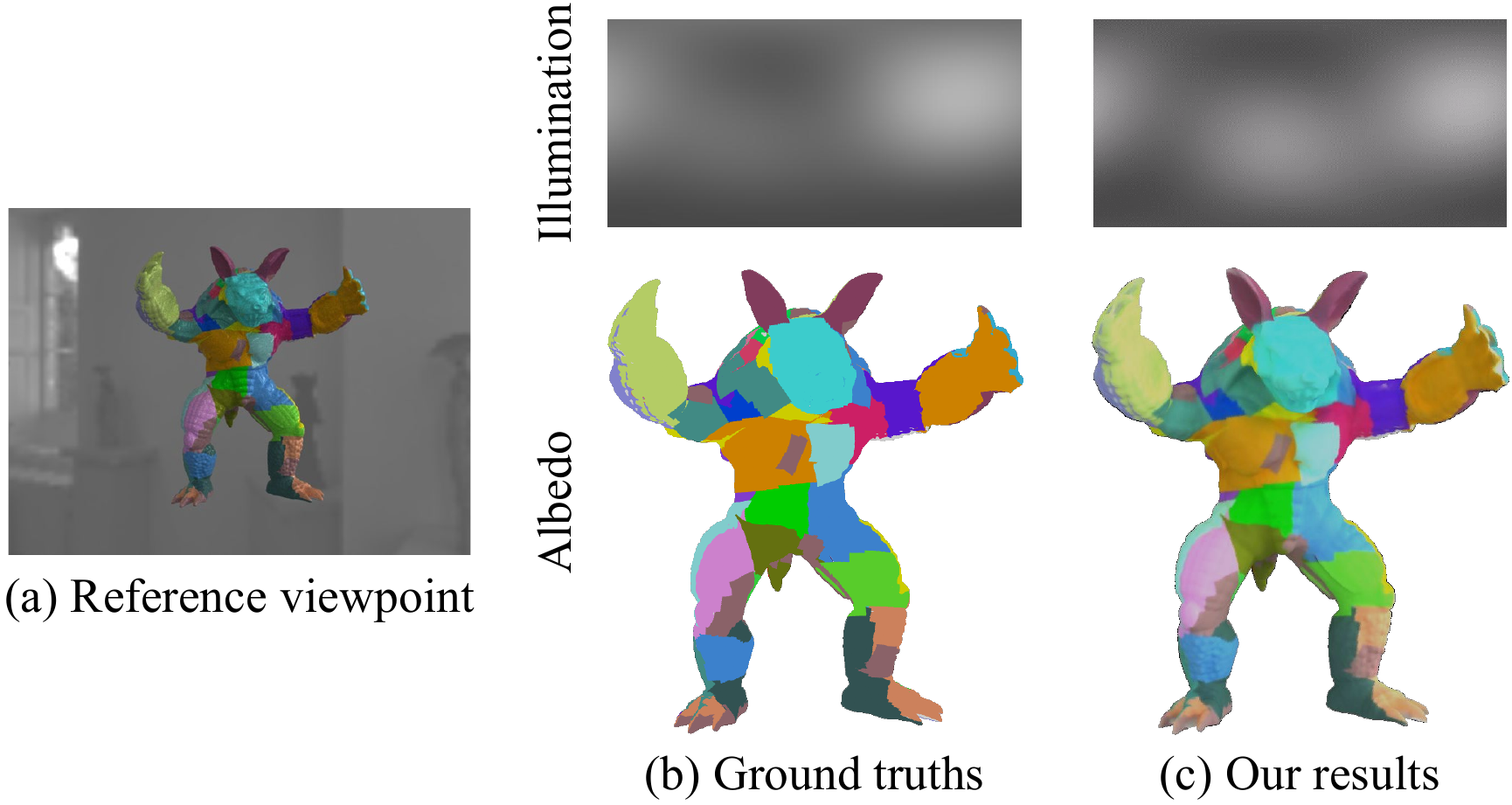}
   \caption{Estimated albedo and illumination: (a) Reference viewpoint; (b) Ground-truth albedo and ground-truth second-order spherical harmonics approximation of the environment map shown in Fig. 7; (c) Our estimation results.}
   \label{fig:illumAndAlbedo}
\end{figure}

Figure~\ref{fig:illumAndAlbedo} shows the examples of our illumination and albedo results. For the illumination, since we adopt the second-order spherical harmonics model, which represents the low-frequency part of the illumination, the high-frequency details of the original environment map are lost. Even though, from the comparison of our estimation result and the ground-truth second-order spherical harmonics approximation, we can confirm that our method obtains a reasonable illumination result.
For the albedo, a reasonable result is also derived, where the overall result is close to the ground truth, though albedos in some regions show the differences from the ground truths. These differences could be derived that the change of albedo and shading are not perfectly distinguished.

\subsubsection{Evaluation using different materials}
\label{sec:material}
\begin{figure*}[t!]
   \centering
   \includegraphics[trim={0cm 0cm 0cm 0cm}, width=1.0\linewidth]{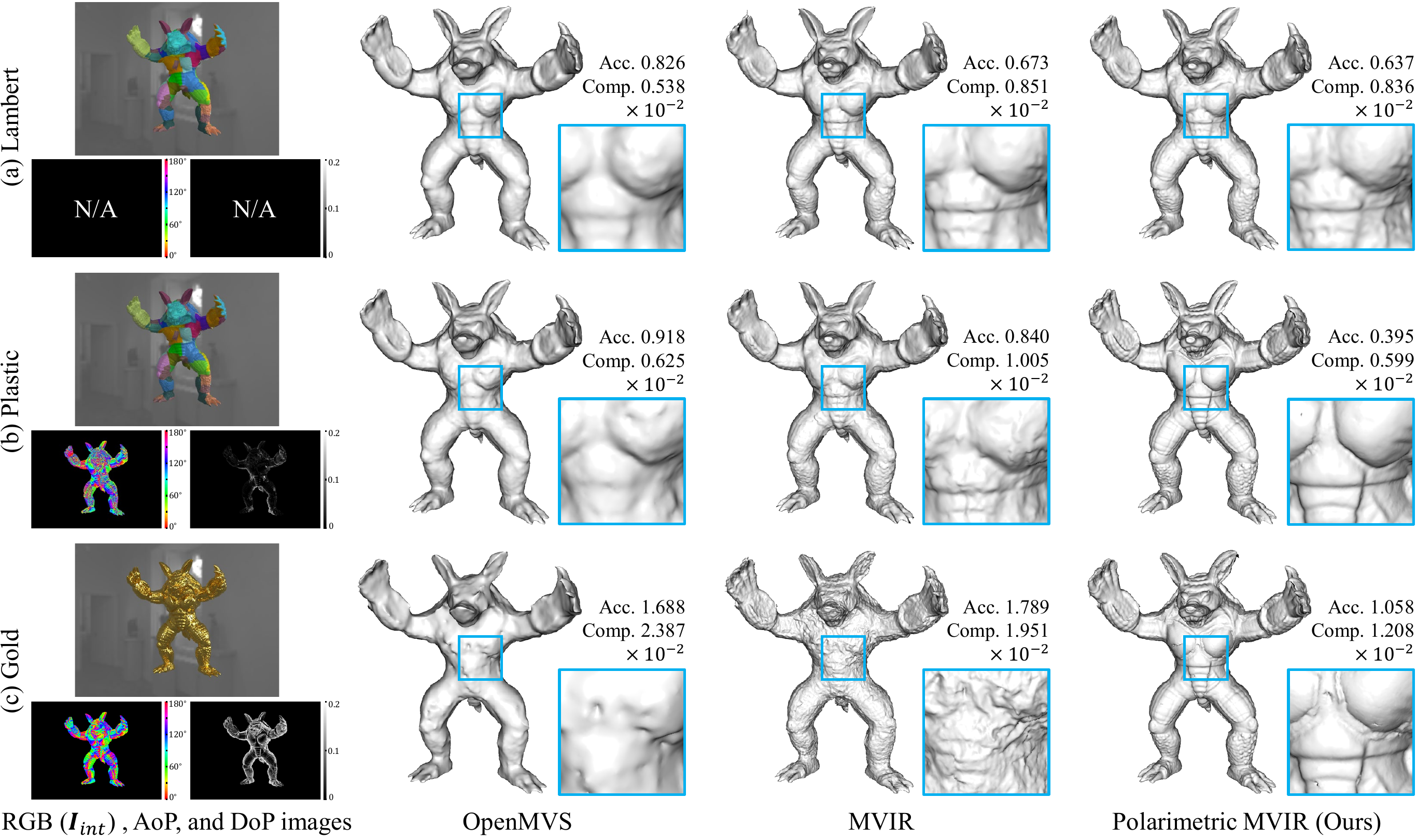}
   \caption{Visual and numerical comparisons of reconstructed 3D models for three different materials: (a) Lambert (unpolarized reflection), (b) Plastic, and (c) Gold.}
   \label{fig:changeSpecular1}
\end{figure*}

In real-world situations, the strengths of DoPs are different under various conditions, such as lighting conditions and object materials. The existing of $\pi$- and $\pi/2$-ambiguities also varies depending on these conditions, as explained in Section~\ref{sec:ambiguities}. For example, if the materials are different, they will result in different AoP and DoP observations at the pixel by pixel even under identical camera and illumination settings. The AoP and the DoP images in Fig.~\ref{fig:changeSpecular1} show such examples, where the images were synthesized for three different materials~(Lambert, plastic, and gold) by using the same camera and illumination settings as Fig.~\ref{fig:setting}(a). In this subsection, we show experimental results using these three materials to demonstrate the robustness of our method against the polarimetric ambiguities. 

The visual and numerical comparisons for the three materials are shown in Fig.~\ref{fig:changeSpecular1}, where we compared MVIR~\cite{kim2016multi}, our Polarimetric MVIR, and OpenMVS~\cite{OpenMVS} as the method to generate the initial model for MVIR and Polarimetric MVIR. 
Figure~\ref{fig:changeSpecular1}(a) shows the case with Lambert reflection, meaning that no polarimetric information is included in the input images. This is an ideal condition for MVIR and there is no advantage for our method using polarimetric information. More specifically, in this case, our method reduces to the existing MVIR, since the polarimetric term is not applied, i.e. the weight for this term becomes zero, due to the absence of DoPs. From the results of Fig.~\ref{fig:changeSpecular1}(a), we can confirm that, even in the case without the polarization, our method can refine the initial OpenMVS result and produce almost the same result as MVIR, as we expected.

Figures~\ref{fig:changeSpecular1}(b) and~\ref{fig:changeSpecular1}(c) show the results for the plastic and the gold materials, respectively.
Comparing the DoP images in Figs.~\ref{fig:changeSpecular1}(b) and~\ref{fig:changeSpecular1}(c), we can see that the gold material generally shows higher DoPs. At the same time, it shows stronger specular highlights, as we can observe from the comparison of the RGB images in Figs.~\ref{fig:changeSpecular1}(b) and~\ref{fig:changeSpecular1}(c), due to stronger polarized specular reflections. From the results of Figs.~\ref{fig:changeSpecular1}(b) and~\ref{fig:changeSpecular1}(c), we can confirm that our method significantly refines the initial OpenMVS results for both materials, while MVIR fails to improve the accuracy for the gold because of the existence of strong specular highlights. The successes of the shape refinement for both materials demonstrate the robustness of our method against polarimetric ambiguities under mixed polarization reflections, which can be observed as different AoP observations in Figs.~\ref{fig:changeSpecular1}(b) and~\ref{fig:changeSpecular1}(c).

Regarding the final 3D model quality reconstructed by our method, the plastic shows a better result than the gold. This is because that the initial model reconstructed by OpenMVS is worse for the gold due to the specular highlights, as can be seen in the results of OpenMVS in Figs.~\ref{fig:changeSpecular1}(b) and~\ref{fig:changeSpecular1}(c).
However, since the gold shows higher DoPs, our method shows a larger improvement for the gold, i.e. the accuracy improvement from OpenMVS is $0.630 \times 10^{-2}$ for the gold, while it is $0.523 \times 10^{-2}$ for the plastic. This demonstrates the effectiveness of our polarimetric term with the DoP weighting. In the next subsection, we numerically evaluate the effectiveness of each proposed component in more details as an ablation study.

\subsubsection{Ablation study}
\label{sec:ablation}
We evaluated the effectiveness of each proposed component using the plastic material by comparing the four methods shown in Table~\ref{table:ablation}.
The first method is a baseline MVIR method that does not apply any polarimetric information. This method was implemented by changing the parameter $\tau_1$ of Eq.~(\ref{eq:costFunction}) to $0$ to eliminate the effect of the polarimetric term. 
The second, the third, and the fourth methods incrementally applied the polarimetric term, the DoP weight for the polarimetric term, and the ${\bm I}_{min}$ input to the cost optimization, to evaluate the effectiveness of each component one by one.
In Table~\ref{table:ablation}, the albedo was evaluated as the average pixel value RMSE between the estimated albedo maps and the ground-truth albedo maps for all the camera views. The illumination was also evaluated as the average pixel value RMSE between the estimated cube maps and the ground-truth cube maps for all the camera views. 

\begin{table}[t!]
   \centering
   \renewcommand\arraystretch{1.2}
   \setlength\aboverulesep{0pt}\setlength\belowrulesep{0pt}
   \setcellgapes{1.5pt}\makegapedcells
   \caption{Results of the ablation study}
   \label{table:ablation}
\begin{tabular}{c|ll|l|l|l|l}
   \toprule
   \multicolumn{3}{c|}{Polarimetric term} & \multicolumn{1}{p{4.5em}|}{} & \multicolumn{1}{p{4.5em}|}{\checkmark} & \multicolumn{1}{p{4.5em}|}{\checkmark} & \multicolumn{1}{p{4.5em}}{\checkmark} \\
   \midrule
   \multicolumn{3}{c|}{DoP weight} &       &       & \multicolumn{1}{p{4.5em}|}{\checkmark} & \multicolumn{1}{p{4.5em}}{\checkmark} \\
   \midrule
   \multicolumn{3}{c|}{Input ${\bm I}_{min}$} &       &       &       & \multicolumn{1}{p{4.5em}}{\checkmark} \\
   \midrule
   \midrule
   \multirow{4}[8]{*}{\begin{sideways}Armadillo\end{sideways}} & \multicolumn{1}{c|}{\multirow{2}[4]{*}{\begin{sideways}Shape\end{sideways}}} & \multicolumn{1}{p{8em}|}{Acc.($\times10^{-2}$)} & 0.840  & 0.537  & \textcolor{red}{\textbf{0.394 }} & 0.395  \\
   \cmidrule{3-7}      & \multicolumn{1}{c|}{} & \multicolumn{1}{p{8em}|}{Comp.($\times10^{-2}$)} & 1.005  & 0.739  & \textcolor{red}{\textbf{0.598 }} & 0.599  \\
   \cmidrule{2-7}      & \multicolumn{2}{l|}{Albedo} & 0.150  & 0.144  & 0.158  & \textcolor{red}{\textbf{0.135 }} \\
   \cmidrule{2-7}      & \multicolumn{2}{l|}{Illumination} & \textcolor{red}{\textbf{0.505 }} & 0.538  & 0.559  & 0.513  \\
   \midrule
   \multirow{4}[8]{*}{\begin{sideways}Bunny\end{sideways}} & \multicolumn{1}{c|}{\multirow{2}[4]{*}{\begin{sideways}Shape\end{sideways}}} & \multicolumn{1}{p{8em}|}{Acc.($\times10^{-2}$)} & 1.646  & 1.154  & 0.890  & \textcolor{red}{\textbf{0.880 }} \\
   \cmidrule{3-7}      & \multicolumn{1}{c|}{} & \multicolumn{1}{p{8em}|}{Comp.($\times10^{-2}$)} & 1.876  & 1.435  & 1.134  & \textcolor{red}{\textbf{1.126 }} \\
   \cmidrule{2-7}      & \multicolumn{2}{l|}{Albedo} & 0.259  & 0.227  & 0.276  & \textcolor{red}{\textbf{0.225 }} \\
   \cmidrule{2-7}      & \multicolumn{2}{l|}{Illumination} & 0.561  & 0.619  & 0.651  & \textcolor{red}{\textbf{0.530 }} \\
   \midrule
   \multirow{4}[8]{*}{\begin{sideways}Dragon\end{sideways}} & \multicolumn{1}{c|}{\multirow{2}[4]{*}{\begin{sideways}Shape\end{sideways}}} & \multicolumn{1}{p{8em}|}{Acc.($\times10^{-2}$)} & 1.274  & 0.942  & 0.746  & \textcolor{red}{\textbf{0.738 }} \\
   \cmidrule{3-7}      & \multicolumn{1}{c|}{} & \multicolumn{1}{p{8em}|}{Comp.($\times10^{-2}$)} & 2.072  & 1.796  & 1.566  & \textcolor{red}{\textbf{1.561 }} \\
   \cmidrule{2-7}      & \multicolumn{2}{l|}{Albedo} & \textcolor{red}{\textbf{0.177 }} & 0.215  & 0.227  & 0.216  \\
   \cmidrule{2-7}      & \multicolumn{2}{l|}{Illumination} & 0.516  & 0.540  & 0.553  & \textcolor{red}{\textbf{0.513 }} \\
   \midrule
   \midrule
   \multirow{4}[8]{*}{\begin{sideways}Average\end{sideways}} & \multicolumn{1}{c|}{\multirow{2}[4]{*}{\begin{sideways}Shape\end{sideways}}} & \multicolumn{1}{p{8em}|}{Acc.($\times10^{-2}$)} & 1.253  & 0.878  & 0.676  & \textcolor{red}{\textbf{0.671 }} \\
   \cmidrule{3-7}      & \multicolumn{1}{c|}{} & \multicolumn{1}{p{8em}|}{Comp.($\times10^{-2}$)} & 1.651  & 1.323  & 1.099  & \textcolor{red}{\textbf{1.095 }} \\
   \cmidrule{2-7}      & \multicolumn{2}{l|}{Albedo} & 0.196  & 0.195  & 0.221  & \textcolor{red}{\textbf{0.192 }} \\
   \cmidrule{2-7}      & \multicolumn{2}{l|}{Illumination} & 0.527  & 0.566  & 0.588  & \textcolor{red}{\textbf{0.519 }} \\
   \bottomrule
   \end{tabular}%   
 \end{table}%

The comparison of the results of the first and the second methods validates that the existence of the polarimetric term is very effective in refining the 3D shape, resulting in much better accuracy and completeness. The integration of the DoP weight to the polarimetric term (the third method) further improves the performance in the shape estimation, where both the accuracy and the completeness become quite better. We can also confirm that our final proposed method with the ${\bm I}_{min}$ input (the fourth method) provides the best average results in all the shape, the albedo, and the illumination evaluations.

From the results of Table~\ref{table:ablation}, we can observe that the performance improvements from the baseline MVIR method (the first method) for the albedo and the illumination are relatively small compared with the improvement for the shape.
This is because our polarimetric term is designed to exploit the physical relationship between the polarization and the surface normal for shape refinement, but not designed to directly improve the illumination and the albedo, since no physical relationship between them and the polarization is considered. Even though, the improvement of the shape contributes to the slight improvements of the illumination and the albedo on average in terms of the global optimization. However, we also observe that this is not always the case (e.g. illumination of Almadillo and albedo of Dragon), which may be because of the imperfection of the reflection model (e.g. not considering shadows and inter-reflections), as well as the imperfection of the illumination model (approximated by the second-order spherical harmonics).

\subsection{Comparison using real data}

\begin{figure*}[t!]
   \centering
   \includegraphics[trim={0cm 0cm 0cm 0cm}, width=1.0\linewidth]{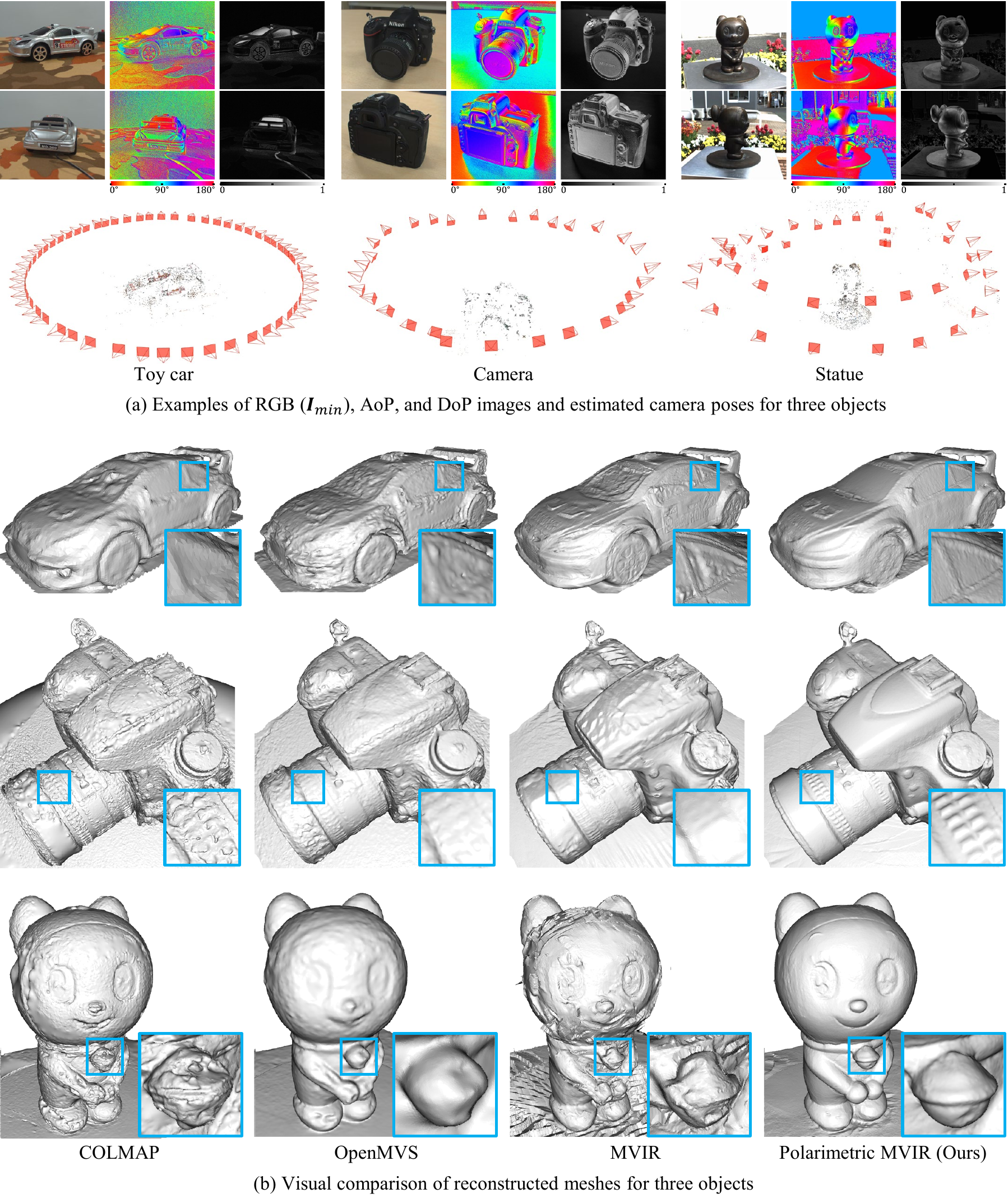}\\
    \vspace{-2mm}
   \caption{Visual comparison using real data.}
   \label{fig:car}
\end{figure*}

\begin{figure*}[t!]
   \centering
   \includegraphics[trim={0cm 0cm 0cm 0cm}, width=0.9\linewidth]{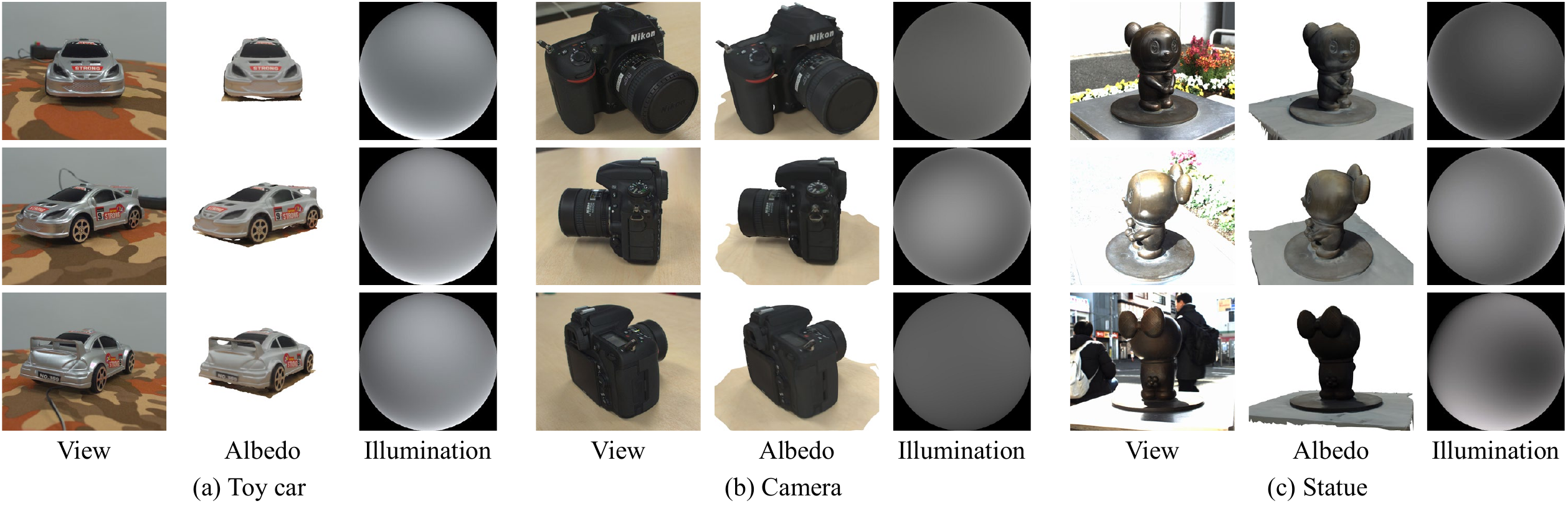}
   \caption{Examples of estimated albedos and illuminations for three real objects.}
   \label{fig:realScene_illumAndAlbedo}
\end{figure*}

Figure~\ref{fig:car} shows the visual comparison of the reconstructed 3D models using real images of a toy car (56 views) and a camera (31 views), which were captured under a normal lighting condition in the office using fluorescent light on the ceiling, and those of a statue (43 views), which were captured under outdoor daylight. We captured the polarization images using Lucid PHX050S-Q camera~\cite{Lucid}. Due to the space limitation, we here compared our method with COLMAP and OpenMVS, which provide the best accuracy and the best completeness, respectively, among the existing methods compare in Table~\ref{table:evaluation}. We also compare MVIR as the baseline MVIR method without using the polarimetric information. The visual comparison of all the methods in Table~\ref{table:evaluation} can be seen in our supplementary material.

The results of Fig.~\ref{fig:car} show that COLMAP can reconstruct fine details in relatively well-textured regions (e.g. the details of the camera lens), while it fails in texture-less regions (e.g. the front window of the car). OpenMVS can better reconstruct the overall shapes owing to the denser points, though some fine details are lost. MVIR performs well except for dark regions, where the shading information is limited (e.g. the top of the camera and the surface of the statue). In contrast, our method can recover finer details and clearly improve the reconstructed 3D model quality by exploiting both photometric and polarimetric information, especially in the regions such as the front body and the window of the toy car and the overall surfaces of the camera and the statue.
Example results of albedos and illuminations by our method are shown in Fig.~\ref{fig:realScene_illumAndAlbedo}. Compared to the results of the toy car and the camera captured indoor, the results of the  statue captured outdoor show more apparent illumination directions, which reasonably represents the directions from the sunlight.

\begin{figure*}[t!]
   \centering
   \vspace{5mm}
   \includegraphics[trim={0cm 0cm 0cm 0cm}, angle=0, width=1.0\linewidth]{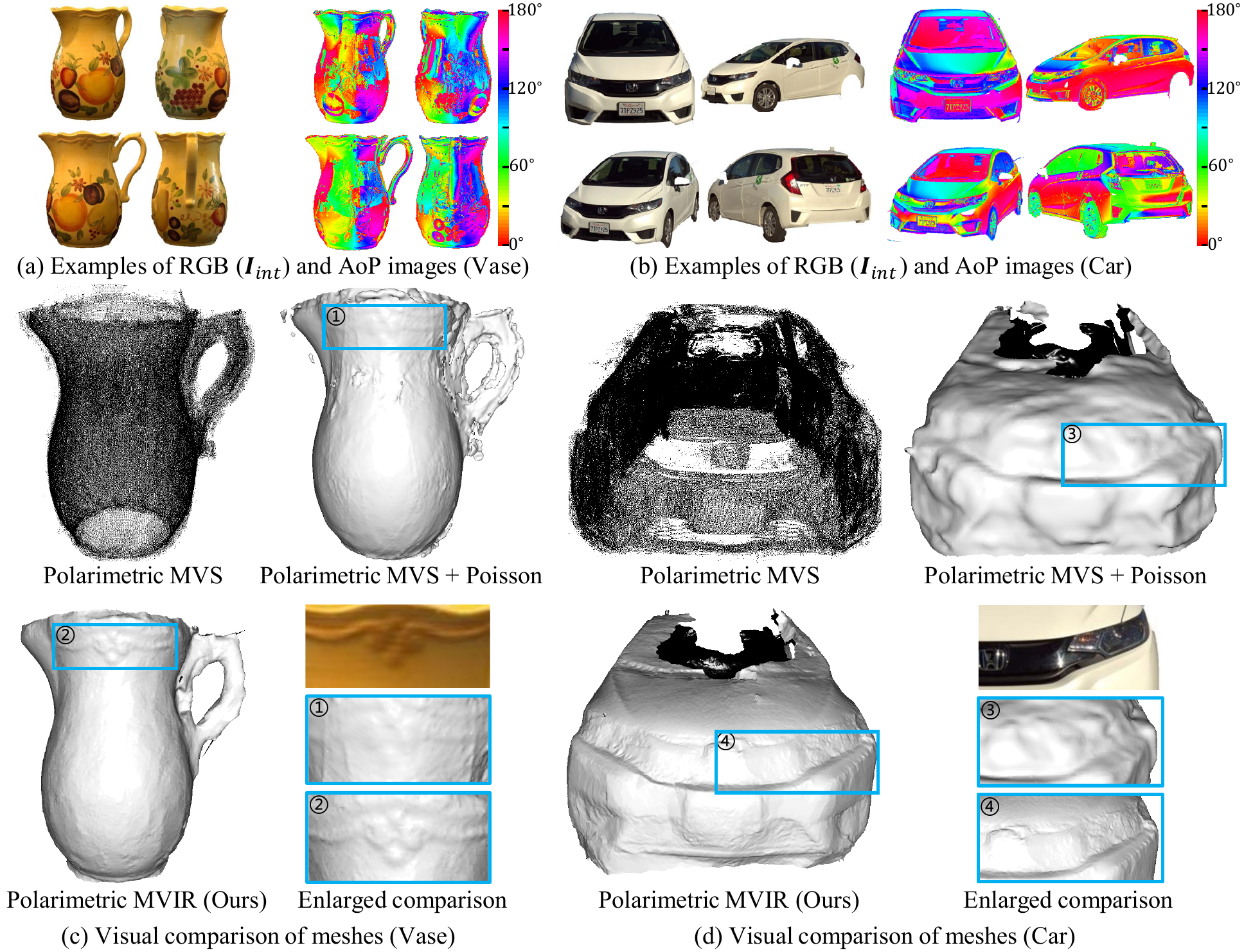}
   \caption{Refinement results for Polarimetric MVS~\cite{cui2017polarimetric} using the data provided by the authors.}
   \label{fig:vase}
\end{figure*}

\begin{figure}[t!]
   \centering
   \includegraphics[trim={0cm 0cm 0cm 0cm}, width=1.0\linewidth]{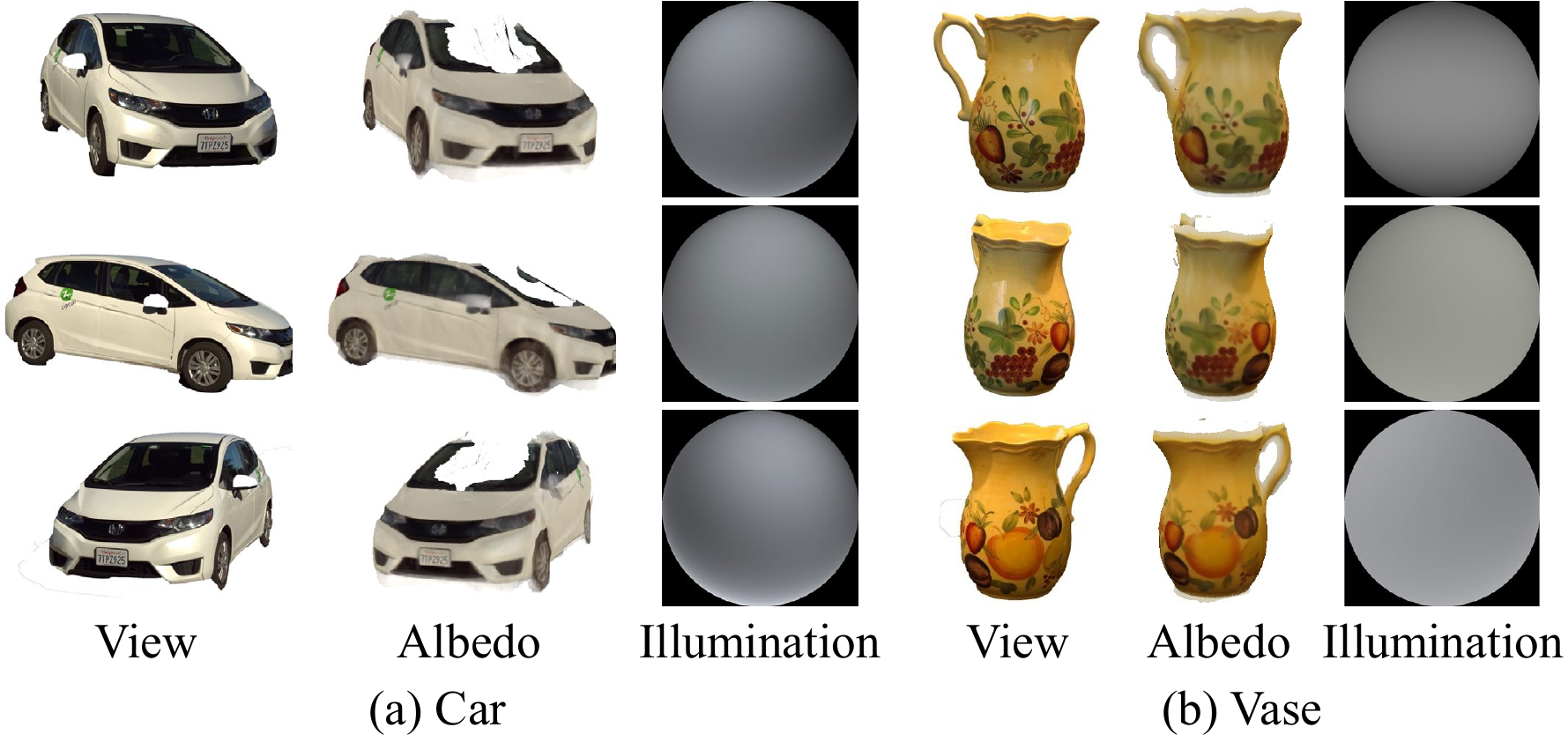}
   \caption{Examples of estimated albedos and illuminations for the real-world data provided by the authors of Polarimetric MVS~\cite{cui2017polarimetric}.}
   \label{fig:polarimetricMVS_illumAndAlbedo}
\end{figure}

\subsection{Refinement for Polarimetric MVS~\cite{cui2017polarimetric}}

Our Polarimetric MVIR adopts the result of an existing MVS method for initial surface generation. Since Polarimetric MVS~\cite{cui2017polarimetric} has succeeded in reconstructing dense points even for texture-less regions by making better use of polarimetric information, combining our Polarimetric MVIR with Polarimetric MVS has great potential to further improve the quality of reconstructed 3D models. 

To confirm this, we conducted the experiments using the data provided by the authors of Polarimetric MVS~\cite{cui2017polarimetric}.
We applied Screened Poisson Surface Reconstruction to the point cloud result obtained by Polarimetric MVS to generate initial surface. Then, we refined the initial surface using the provided camera poses, and RGB~(${\bm I}_{int}$) and AoP images from 36 viewpoints, where we did not apply RGB~(${\bm I}_{min}$) images and the DoP weighting for our method because necessary DoP images have not been provided.

Figure~\ref{fig:vase} shows the refinement results of two provided objects (vase and car). 
Polarimetric MVS can provide dense point clouds, even for texture-less regions such as the body of the car, by exploiting polarimetric information. However, there are still some outliers, which could be derived from the noise of AoP measurements and incorrect disambiguation, and resultant surfaces are rippling. These artifacts are alleviated in our Polarimetric MVIR by solving the ambiguity problem as a global cost optimization. Moreover, we can see that finer details are reconstructed using photometric shading information, which Polarimetric MVS does not exploit.
Example results of albedos and illuminations by our method are shown in Fig.~\ref{fig:polarimetricMVS_illumAndAlbedo}, where we can visually see that reasonable albedo and illumination results are derived.

\section{Conclusions}
In this paper, we have proposed Polarimetric MVIR, which can reconstruct a high-quality 3D model by optimizing multi-view photometric rendering errors and polarimetric errors. Polarimetric MVIR resolves the $\pi$- and $\pi/2$-ambiguities as an optimization problem, which makes the method fully passive and applicable to various materials. 
Experimental results have demonstrated that Polarimetric MVIR is robust to ambiguities, and generates more detailed 3D models compared with existing state-of-the-art multi-view reconstruction methods.
Our Polarimetric MVIR has a limitation that it requires a reasonably good initial shape for its global optimization, which would encourage us to develop a more robust initial shape estimation method.

\ifCLASSOPTIONcompsoc
   \section*{Acknowledgments}
\else
   \section*{Acknowledgment}
\fi

This work was partly supported by JSPS KAKENHI Grant Number 17H00744 and 21K17762. The authors would like to thank Dr. Zhaopeng Cui for sharing the data of Polarimetric MVS.

\ifCLASSOPTIONcaptionsoff
   \newpage
\fi

\bibliographystyle{IEEEtran}
\bibliography{egbib}

\begin{IEEEbiography}[{\includegraphics[width=1in,height=1.25in,clip,keepaspectratio]{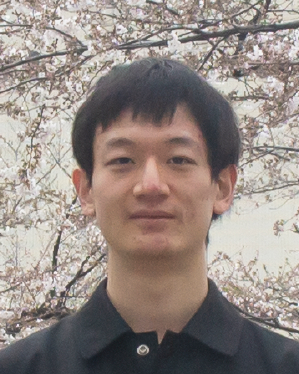}}]{Jinyu Zhao}
   received the B.Eng. degree from the Department of Automation, Wuhan University, Wuhan, China, in 2018, and the M.Eng. degree from the Department of Systems and Control Engineering, Tokyo Institute of Technology, Tokyo, Japan, in 2020. Currently he is a PhD student in the Department of Systems and Control Engineering, Tokyo Institute of Technology. His research interests include polarization imaging and computer vision.
\end{IEEEbiography}

\begin{IEEEbiography}[{\includegraphics[width=1in,height=1.25in,clip,keepaspectratio]{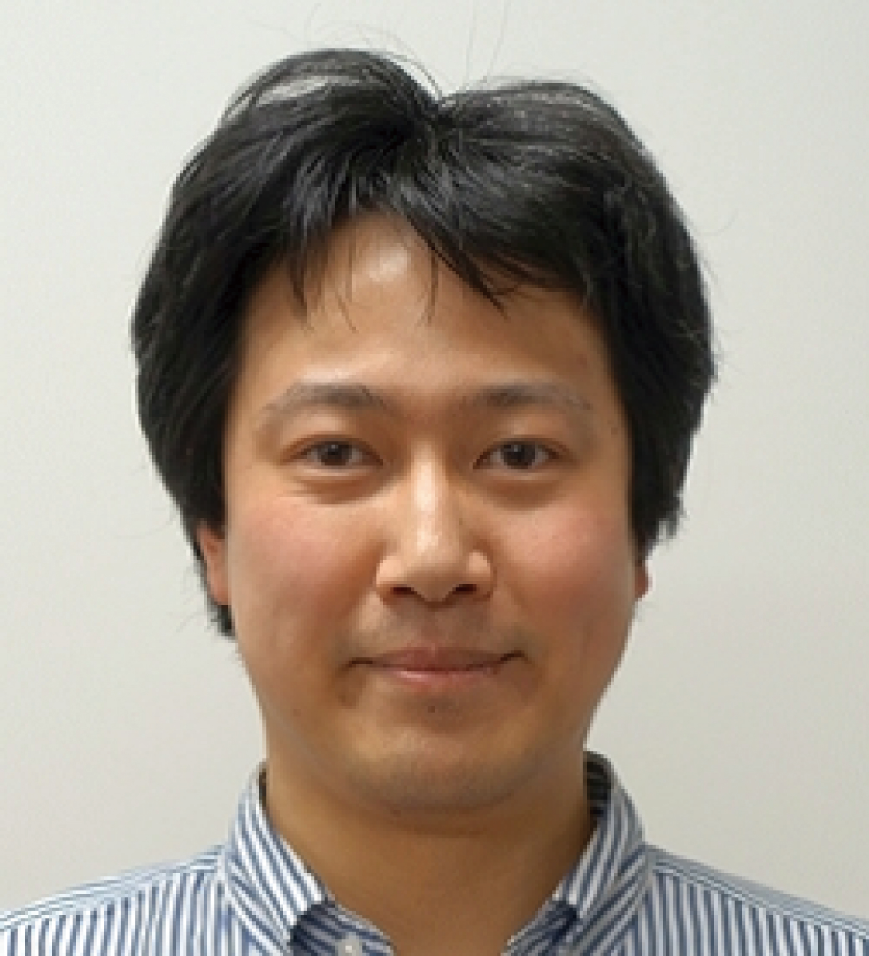}}]{Yusuke Monno}
received the B.E., M.E., and Ph.D degrees from Tokyo Institute of Technology, Tokyo, Japan, in 2010, 2011, and 2014, respectively. From Nov. 2013 to Mar. 2014, he joined the Image and Visual Representation Group at École Polytechnique Fédérale de Lausanne as a research internship student. He is currently an assistant professor with the Department of Systems and Control Engineering, School of Engineering, Tokyo Institute of Technology. His research interests are in both theoretical and practical aspects of image processing, computer vision, and biomedical engineering.
\end{IEEEbiography}
\vspace{-140mm}

\begin{IEEEbiography}[{\includegraphics[width=1in,height=1.25in,clip,keepaspectratio]{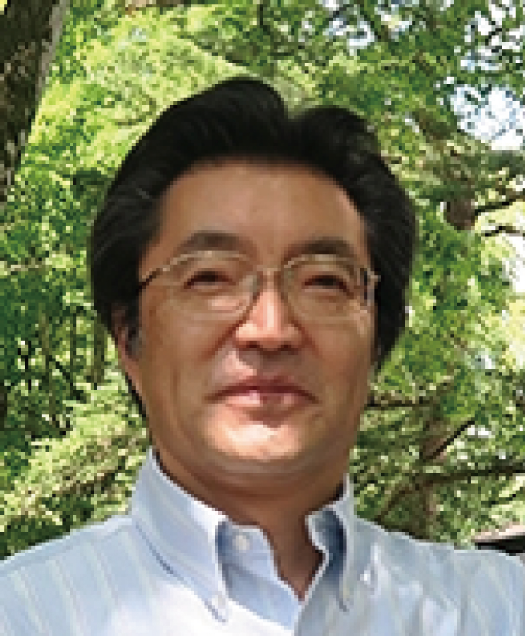}}]{Masatoshi Okutomi}
received the B.Eng. degree from the Department of Mathematical Engineering and Information Physics, the University of Tokyo, Tokyo, Japan, in 1981, and the M.Eng. degree from the Department of Control Engineering, Tokyo Institute of Technology, Tokyo, in 1983. He joined the Canon Research Center, Canon Inc., Tokyo, in 1983. From 1987 to 1990, he was a Visiting Research Scientist with the School of Computer Science, Carnegie Mellon University, Pittsburgh, PA, USA. He received the Dr.Eng. degree from Tokyo Institute of Technology, in 1993, for his research on stereo vision. Since 1994, he has been with Tokyo Institute of Technology, where he is currently a Professor with the Department of Systems and Control Engineering, the School of Engineering. 
\end{IEEEbiography}

\end{document}